%% file: sample-sigconf.tex
\documentclass[acmsmall]{acmart}
\usepackage{subcaption}
\usepackage{booktabs}
\usepackage{multirow}
\usepackage{array}
\usepackage{makecell}
\usepackage{xcolor}
\usepackage{bm}        
\usepackage{amsmath}
\usepackage{tcolorbox}
\AtBeginDocument{%
  }

\setcopyright{cc}
\setcctype{by}
\acmJournal{PACMMOD}
\acmYear{2026} \acmVolume{4} \acmNumber{1 (SIGMOD)} \acmArticle{41} \acmMonth{2} \acmPrice{}

\acmDOI{10.1145/3786655}

\begin{document}

\title{From Prefix Cache to Fusion RAG Cache: Accelerating LLM Inference in Retrieval-Augmented Generation}

\author{Jiahao Wang}
\authornote{These authors contributed equally to this research.}
\email{202241050020@hdu.edu.cn}
\affiliation{%
  \institution{Hangzhou Dianzi University}
  \city{Hangzhou}
  \country{China}
}
\affiliation{%
  \institution{Approaching.AI}
  \country{China}
}

\author{Weiyu Xie}
\authornotemark[1]
\email{xwy21@mails.tsinghua.edu.cn}
\affiliation{%
  \institution{Tsinghua University}
  \city{Beijing}
  \country{China}
}

\author{Mingxing Zhang}
\authornote{Mingxing Zhang is the corresponding author.}
\email{zhang\_mingxing@mail.tsinghua.edu.cn}
\affiliation{%
  \institution{Tsinghua University}
  \city{Beijing}
  \country{China}
}

\author{Boxing Zhang}
\email{zhangbx24@mails.tsinghua.edu.cn}
\affiliation{%
  \institution{Tsinghua University}
  \city{Beijing}
  \country{China}
}

\author{Jianwei Dong}
\email{dongjw24@mails.tsinghua.edu.cn}
\affiliation{%
  \institution{Tsinghua University}
  \city{Beijing}
  \country{China}
}

\author{Yuening Zhu}
\email{zyn2003715@163.com}
\affiliation{%
  \institution{Tsinghua University}
  \city{Beijing}
  \country{China}
}

\author{Chen Lin}
\email{lin-c24@mails.tsinghua.edu.cn}
\affiliation{%
  \institution{Tsinghua University}
  \city{Beijing}
  \country{China}
}

\author{Jinqi Tang}
\email{azure@approaching.ai}
\affiliation{%
  \institution{Approaching.AI}
  \country{China}
}

\author{Yaochen Han}
\email{ailililisi@approaching.ai}
\affiliation{%
  \institution{Approaching.AI}
  \country{China}
}

\author{Zhiyuan Ai}
\email{awake@approaching.ai}
\affiliation{%
  \institution{Approaching.AI}
  \country{China}
}

\author{Xianglin Chen}
\email{chenxl6436@outlook.com}
\affiliation{%
  \institution{Approaching.AI}
  \country{China}
}

\author{Yongwei Wu}
\email{wuyw@tsinghua.edu.cn}
\affiliation{%
  \institution{Tsinghua University}
  \city{Beijing}
  \country{China}
}

\author{Congfeng Jiang}
\email{cjiang@hdu.edu.cn}
\affiliation{%
  \institution{Hangzhou Dianzi University}
  \city{Hangzhou}
  \country{China}
}


\renewcommand{\shortauthors}{Wang, Xie et al.}

\begin{abstract}
Retrieval-Augmented Generation enhances Large Language Models by integrating external knowledge, which reduces hallucinations but increases prompt length. This increase leads to higher computational costs and longer Time to First Token. To mitigate this issue, existing solutions aim to reuse the preprocessed KVCache of each retrieved chunk to accelerate RAG. However, the lack of cross-chunk contextual information leads to a significant drop in generation quality, leaving the potential benefits of KVCache reuse largely unfulfilled.

The challenge lies in how to reuse the precomputed KVCache chunk while preserving generation quality. We propose FusionRAG, a novel inference framework that optimizes both the preprocessing and reprocessing stages of RAG. In the offline preprocessing stage, we embed information from other related text chunks into each chunk, while in the online reprocessing stage, we recompute the KVCache for tokens that the model focuses on. As a result, we achieve a better trade-off between generation quality and efficiency.
According to our experiments, FusionRAG significantly improves generation quality at the same recomputation ratio compared to previous state-of-the-art solutions. By recomputing fewer than 15\% of the tokens, FusionRAG achieves up to 70\% higher normalized-F1 scores than baselines and reduces TTFT by 2.66-9.39$\times$ compared to Full Attention.
\end{abstract}




\maketitle

\section{Introduction}\label{sec:introduction}

\subsection{Motivation}
\label{sec:motivation}
Retrieval-Augmented Generation(RAG) \cite{lewis2020retrieval, lin2025telerag} is a widely used technique to supplement background knowledge during Large Language Models(LLMs) inference, thereby reducing hallucinations. Recent studies on the inference scaling law of RAG~\cite{Yue2024InferenceSF} show that retrieving more document chunks and performing more retrieval iterations both improve generation results. However, appending retrieved document chunks to the original, relatively shorter question prompt significantly increases the prompt length. This prolonged prompt increases user wait time, measured as Time to First Token (TTFT) and adds computational overhead\cite{Jin2024RAGCacheEK}. For instance, in the Musique dataset, a typical long-context question-answering benchmark with 200 samples, each query processes an average of 17k tokens during the prefill stage, while the subsequent decode stage generates fewer than 5 tokens. The temporal distribution reveals that the prefill stage dominates the overall inference latency, accounting for 95.53\% of the total inference time. This firmly establishes the computational overhead of the prefill stage as the critical bottleneck for end-to-end latency in RAG scenarios. 


To address this issue, previous works \cite{Gao2024CostEfficientLL, kwon_efficient_2023, Lu2024TurboRAGAR, 10.1145/3725273} have explored leveraging KVCache reuse in RAG systems. 
These methods can be categorized into two strategies. 

\textbf{Single-stage methods}, exemplified by Cache-Craft~\cite{10.1145/3725273}, select cache copies from historical user sessions during multi-turn conversations. The selection of cache copies is based on two types of attention characteristics: intra-attention (attention within a chunk) and inter-attention (attention across different chunks).

\begin{figure*}[t]
  \centering
  \includegraphics[trim=0 0 0 0, clip, width=\textwidth]{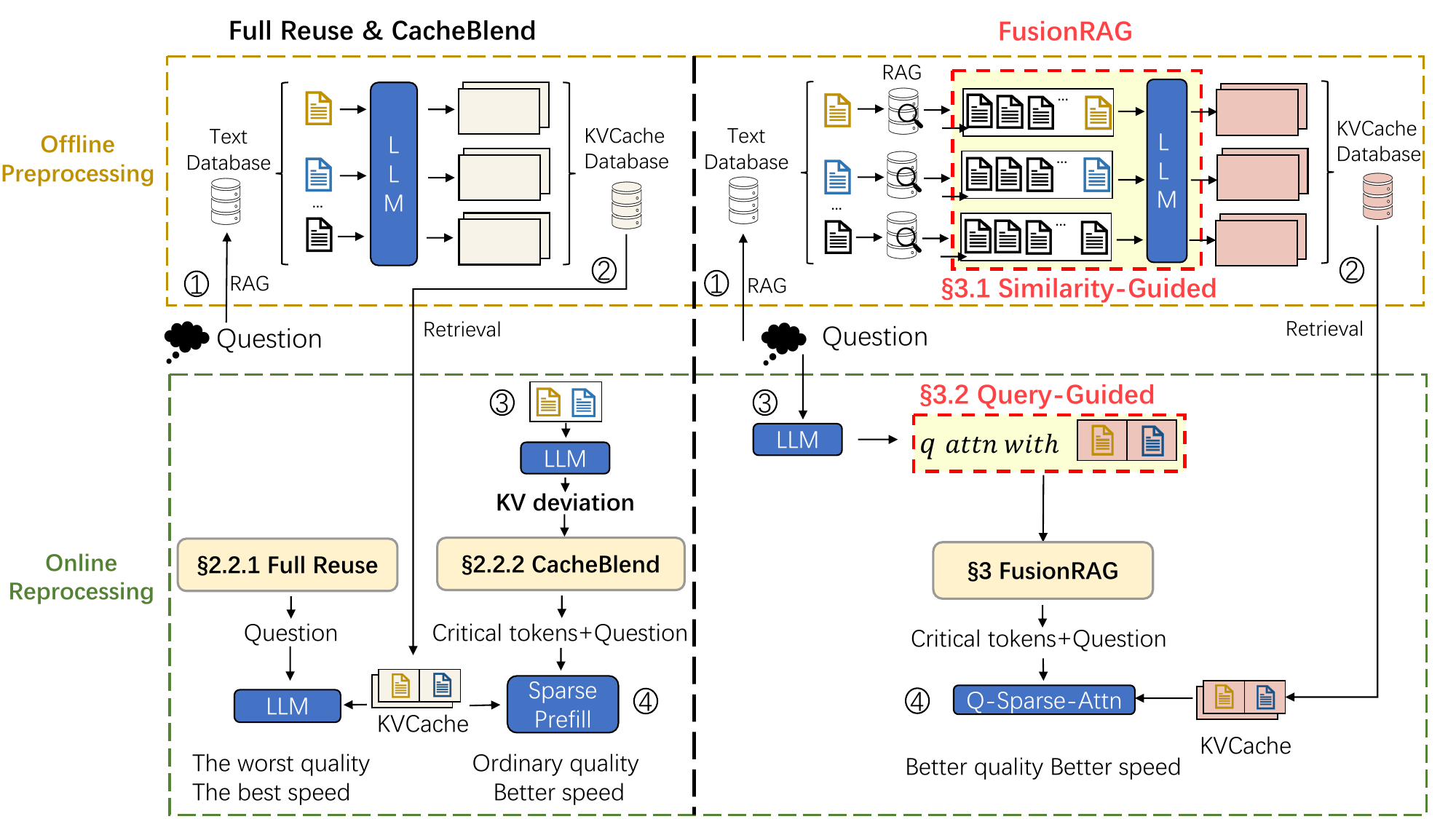}
  \caption{Comparison of Two-stage  methods. \textbf{Left:} Existing methods (Full Reuse and CacheBlend) perform minimal offline preprocessing and rely on online recomputation. \textbf{Right:} FusionRAG introduces Similarity-Guided offline preprocessing (\S\ref{sec:offline_preprocessing}) to precompute cross-attention among similar chunks, and Query-Guided selection (\S\ref{sec:online_reprocessing}) to identify critical tokens online, achieving better quality and speed trade-offs.}
  \label{fig:framework}
\vspace{-10pt}
\end{figure*}

\textbf{Two-stage methods}, represented by TurboRAG~\cite{Lu2024TurboRAGAR} and Cache\-Blend~\cite{yao2025cacheblend}, adopt an \textbf{offline + online} strategy, as shown in Figure \ref{fig:framework}. The offline preprocessing stage is specifically introduced to address the high storage overhead of single-stage methods, which must store multiple KVCache copies for each text chunk. In the offline preprocessing stage, LLMs process each text chunk to generate and save its KVCache. The online stage then bypasses prefill computation by stitching these saved KVCaches together—an approach we call \textbf{\textit{Full Reuse}}. In contrast, \textbf{\textit{Full Attention}} is the standard method of processing the entire prompt online, with no offline preprocessing. 

\textbf{Full Reuse achieves significant TTFT reduction at the cost of generation quality degradation.} This quality decline arises from two sources. First, independently computing KVCache for each chunk produces overlapping position IDs (e.g., [0...l, 0...l, 0...l]) when naively concatenated. TurboRAG~\cite{Lu2024TurboRAGAR} addresses this issue by reordering position IDs into consecutive sequences [0...l, l+1...2l, 2l+1...3l]. Second, even with corrected position IDs, further study~\cite{yao2025cacheblend} reveals a more fundamental problem: the KVCache deviates from Full Attention. This deviation is caused by the lack of cross-attention when chunks are computed independently and results in a noticeable quality decline.

To enhance generation quality, CacheBlend pioneered the selective recomputation framework, where only a small portion of tokens are recomputed based on KV deviation analysis, establishing the foundation for research in this area. Meanwhile, Cache-Craft adopts an alternative strategy, targeting 
tokens with high inter-attention as context-sensitive candidates 
for recomputation.
As illustrated in Figure \ref{fig:framework}, the standard online processing in two-stage methods is replaced with an online reprocessing phase, where critical tokens are identified before the prefill stage. This reprocessing phase aims to minimize the deviation by fully computing the causal attention not only to the user query but also to these critical tokens, thereby restoring generation quality. CacheBlend's core hypothesis is that KVCache deviations are localized to a limited subset of tokens. This implies that recomputing fewer than \textbf{15\%} of tokens can, in principle, recover most of the lost generation quality, offering an optimal trade-off between computational cost and quality.

\begin{figure*}[t]
  \centering
  \begin{subfigure}[t]{0.65\textwidth}
    \centering
    \includegraphics[trim=0 10 0 0, clip, width=\textwidth]{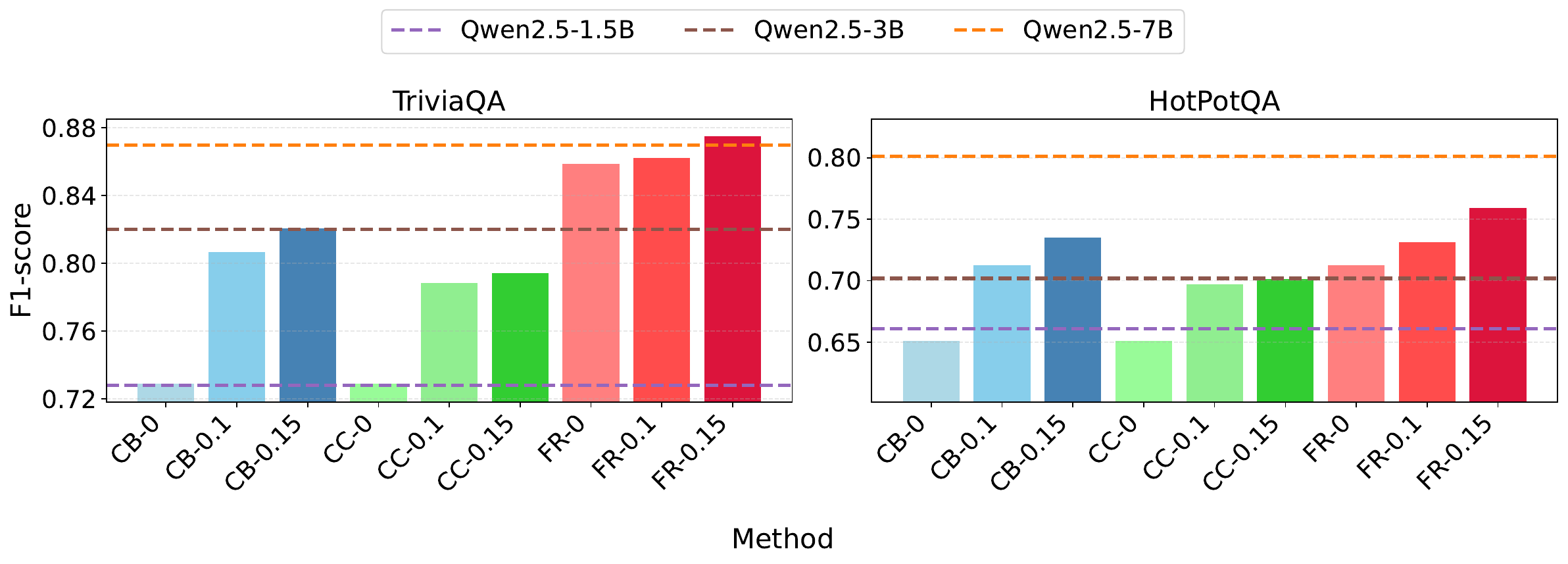}
    \caption{Within the recomputation ratio range of [0, 0.15], we evaluat the generation quality of Qwen2.5-7B under the CB (CacheBlend), CC (Cache-Craft) and FR (FusionRAG), and compared the results with the quality of Qwen2.5 models (1.5B, 3B, and 7B) under the Full Attention.}
    \label{fig:qwen2_series}
  \end{subfigure}
  \hfill
  \begin{subfigure}[t]{0.33\textwidth}
    \centering
    \includegraphics[trim=0 30 170 0, clip, width=\textwidth]{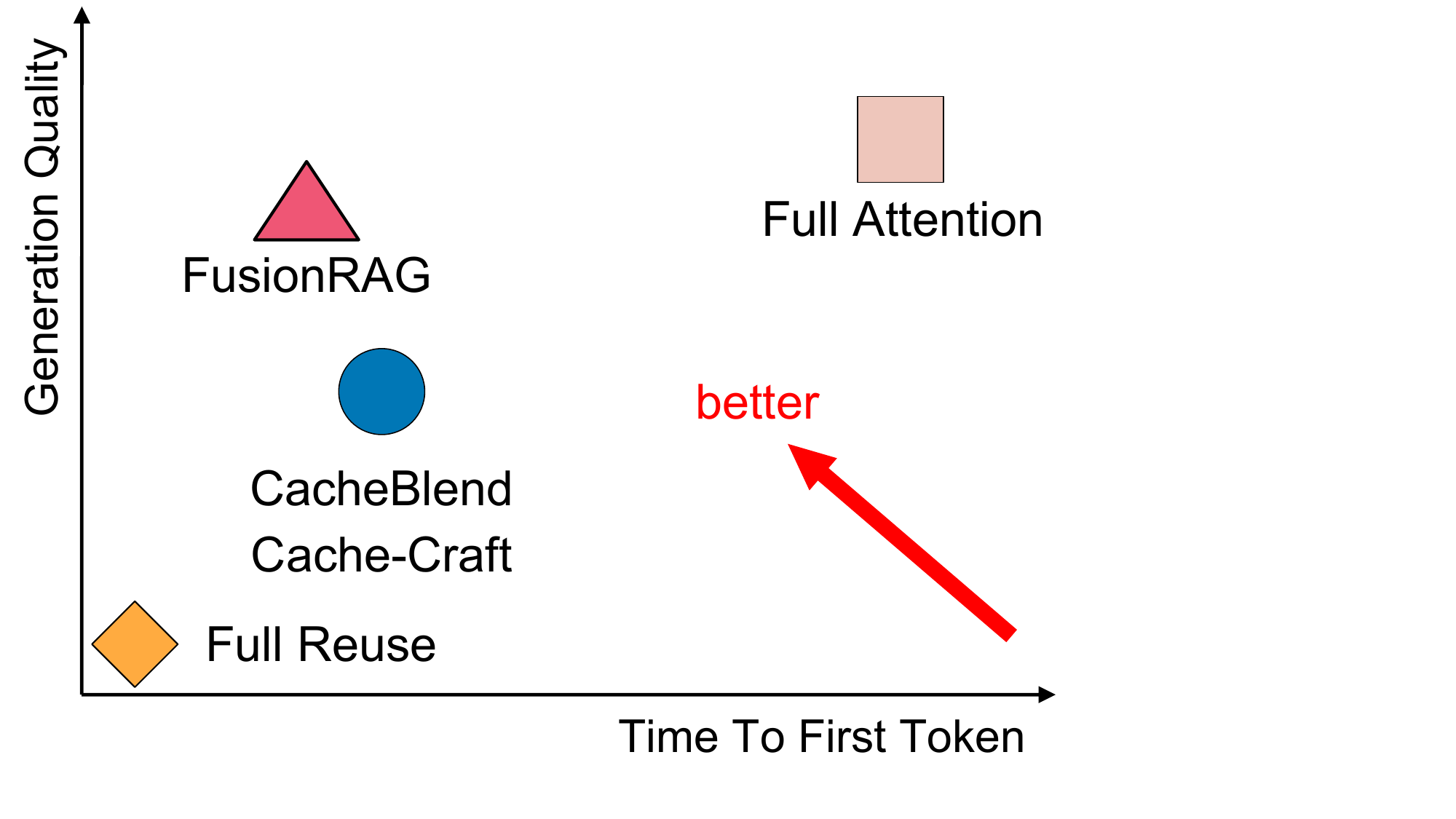}
    \caption{Comparisons of different schemes on performance and generation quality.}
  \end{subfigure}
    \vspace{-5pt}
  \caption{Comparison of the Effectiveness and Efficiency of Existing KVCache Reuse Methods. FusionRAG achieves superior performance in both generation quality and efficiency.}
  \vspace{-10pt}
  \label{fig:methods_comparison}
\end{figure*}

However, there still exists a quality gap between these methods and Full Attention. As shown in Figure \ref{fig:methods_comparison}, we evaluate the generation quality of Qwen2.5-7B-Instruct~\cite{qwen25} with CacheBlend and Cache-Craft mechanisms against smaller Qwen2.5-1.5B-Instruct and Qwen2.5-3B-Instruct models using full attention. As depicted in Figure \ref{fig:methods_comparison}, the generation performance of the Qwen2.5-7B-Instruct model using CacheBlend and Cache-Craft (recomputing only 15\% of tokens) decline sharply, falling to a level comparable to the smaller Qwen2.5-3B-Instruct model operating with full attention. Notably, the 3B model requires less than half of the computational resources for training and inference compared to the 7B model. This discrepancy suggests a critical need for a more effective caching approach that retains both computational efficiency and high-quality generation outcomes.

\begin{figure}[t]
    \centering
    \begin{subfigure}[b]{0.495\linewidth}  
        \includegraphics[width=\linewidth]{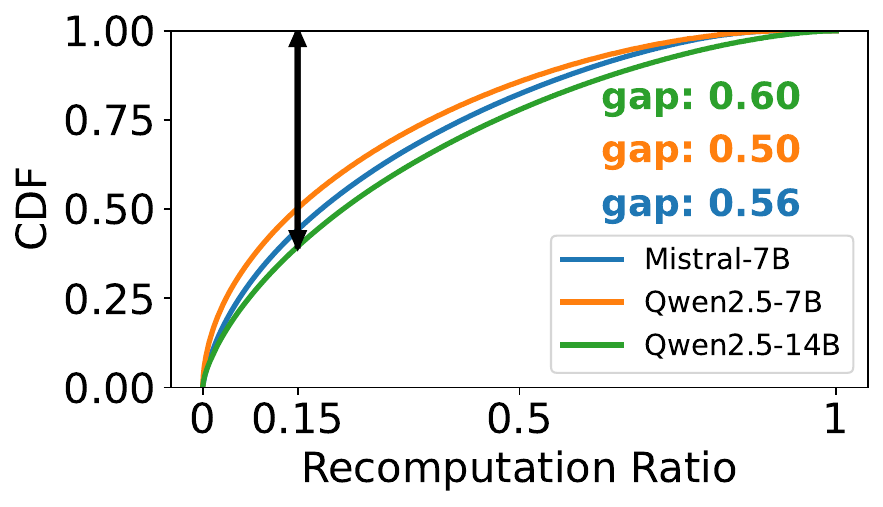}
        \caption{Cumulative probability distribution of KV Deviation.}
        \label{fig:cacheBlend_cdf}
    \end{subfigure}
    \hfill
    \begin{subfigure}[b]{0.495\linewidth}  
        \includegraphics[trim=0 0 0 0, clip, width=\linewidth]{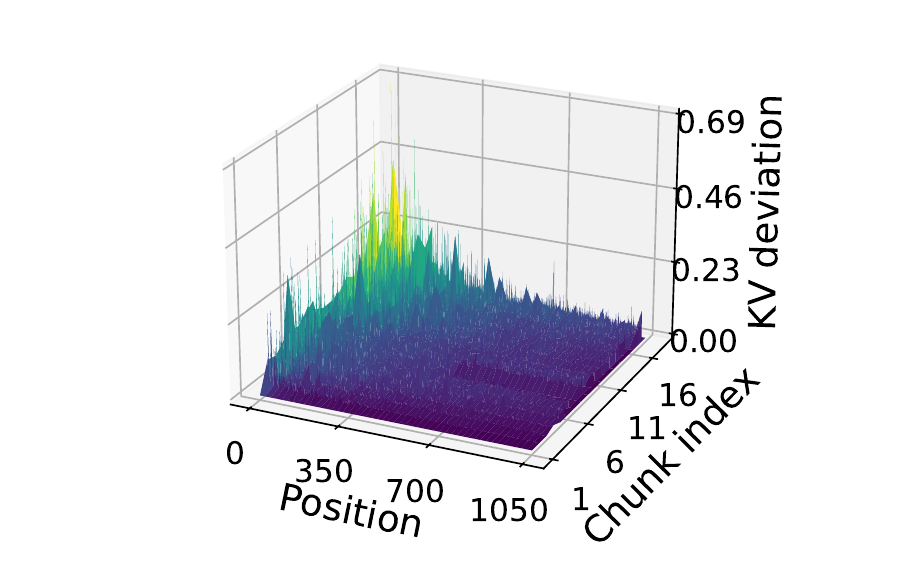}
        \caption{The KV deviation is concentrated in tokens with early positions and higher indices.}
        \label{fig:cacheBlend_deviation_2_layer}
    \end{subfigure}
    \vspace{-15pt}
    \caption{Visualize KV deviation on a Musique question.}
    \vspace{-10pt}
    \label{fig:cacheBlend_defect}
\end{figure}

To achieve a better trade-off between cost and quality, especially at low recomputation ratios, we identify three primary sources of the gap in generation quality. The limitations are as follows:

\textbf{First, existing recomputation methods fail to efficiently restore quality without incurring significant latency.} We analyzed KV deviation for the same question across three different models. As shown in Figure \ref{fig:cacheBlend_cdf}, even after recomputing 15\% of the most critical tokens—identified by a postmortem oracle—over 50\% of the KV deviation still remains in the second layer of the LLM. This demonstrates that a small recomputation budget in the online stage is fundamentally insufficient to compensate for the majority of the KV deviation. Cache-Craft even recommends recomputing over 30\% of tokens to achieve acceptable quality, which confirms that fully restoring quality requires a computational budget so substantial that it significantly increases online latency.

\textbf{Second, existing token selection methods are suboptimal.} For example, Cache\-Blend exhibits a certain bias in selecting tokens for recomputation during the online selection phase: it identifies tokens with high KV deviation in the second layer as critical tokens. As shown in Figure \ref{fig:cacheBlend_deviation_2_layer}, using an example from the Musique dataset, the peak KV deviations in the second layer are predominantly concentrated at the beginning of each chunk. Additionally, tokens located in chunks that are retrieved later generally show higher KV deviations. This phenomenon can be attributed to the fact that each token can only attend to preceding tokens, leading to incomplete contextual information for tokens at the start of a chunk and those appearing in later-retrieved chunks. As a result, these tokens are more likely to exhibit higher KV deviations, making CacheBlend's selection policy less capable of capturing the actual semantic focuses.


\textbf{Third, efficiently scheduling and managing KVCache recomputation presents significant challenges.} For example, the recomputation introduces additional overhead in both cache access and scheduling. This is particularly evident in scenarios involving large-scale models or long input sequences, where frequent cache access (potentially from disk) may offset the inference speedup gained through caching. Additionally, methods like Cache-Craft that maintain multiple KVCache copies for each text chunk across different prefix contexts significantly increase storage overhead.

\subsection{Our Solution}
To address the above issues and achieve a better trade-off between performance and efficiency in the RAG scenario, we optimized both phases of the two-stage RAG framework and proposed a new FusionRAG inference framework, as shown in the right part of Figure \ref{fig:framework}. To tackle the issue of insufficient cross-attention in the online stage, we enhanced the offline preprocessing stage. By leveraging the similarity between the retrieved texts and the positive correlation between their recall probabilities, we performed \textbf{Similarity-Guided Preliminary Cross-attention Preprocessing} during offline preprocessing. As shown in Figure \ref{fig:methods_comparison}, FusionRAG significantly improves generation quality at the same recomputation ratio. In the online reprocessing stage, we introduced a \textbf{Query-Guided Selection} to ensure a more evenly distributed selection of tokens for recomputation. 

To efficiently implement the FusionRAG framework within an LLM serving system, we focus on three key components essential to its integration and execution. First, we introduce Alternative Path, which enhances cache reuse rates by seamlessly integrating with existing prefix-based KVCache reuse. This approach operates without modifying the hash table interface and, crucially, ensures that each text chunk maintains only one replica. Second, we design an asynchronous KVCache and scheduler to address GPU waiting issues caused by I/O blocking during KVCache reads. Finally, we redesign and optimize the sparse attention operator (Q-Sparse-Attn) to make it suitable for FusionRAG's reprocessing stage and support batch decoding.

We implemented FusionRAG on transformers~\cite{wolf2020transformers} and compared it with state-of-the-art KVCache Reuse methods across four QA benchmark datasets and five open-source models.
By recomputing only 15\% of the tokens, we achieved improvements of up to 70\% in normalized-F1-score, surpassing the state-of-the-art methods like CacheBlend and Cache-Craft.

In terms of performance, with 15\% recomputation on a 32K context, Q-Sparse-Attn reduces computation time by approximately up to 77\% compared to the state-of-the-art attention operators FlashAttention \cite{dao2022flashattention} and SDPA from pytorch~\cite{Paszke2019PyTorchAI}. Additionally, compared to Full Attention, FusionRAG reduces TTFT by 2.66-9.39$\times$ when processing 27K contexts in Musique under 15\% KVCache recomputation in end-to-end single question testing. In multi-question testing, FusionRAG achieves a throughput improvement of 1.2-4.3$\times$ compared to the baseline by recomputing 15\% of the KVCache.

\begin{table}[ht]
\caption{Terminology to description cache reuse}
\vspace{-5pt}
\centering
\begin{tabular}{l l}
\toprule
\textbf{Notation} & \textbf{Description} \\
\midrule
$D$, $hd$ & Model's depth and hidden size \\
$X[i]$ & i-th token of the input  \\
$KV[i]$, $KV[:,l]$& i-th token's or $l$-th layer's KVCache \\
$\Delta_{KV}[:,\  l]$ & KV deviation on $l$-th layer\\
\midrule
$C_i, KV_i^{\text{cached}}$ & $i$-th text chunk in knowledge base\\
\quad &  and it's correspondent preprocessed $KV$\\
$C_{\text{top } i}, KV_{\text{top } i}^{\text{cached}}$ & $i$-th retrieved text chunk for a question\\
\quad &  and it's correspondent preprocessed $KV$\\
\midrule
$|\cdot|$ &  The number of tokens in a text \\
$\text{cat}(\cdots)$ &  Concatenate multiple tensors(vectors) \\
\quad & along the token sequence dimension \\
$\text{argTopk}(\cdot)$ & Find the indices  \\
\quad & of the top-$k$ elements in a vector \\
\bottomrule
\end{tabular}
\label{tab1}
\vspace{-5pt}
\end{table}

\section{Preliminary and Problem Definition}\label{sec:back}
\subsection{LLM and Prefix KVCache Reuse}

LLM has attracted significant attention due to its remarkable performance across a wide range of tasks and successful commercialization in real-world applications\cite{techopedia2023, projectpro2023, indatalabs2023, cellstrat2023}. 
It utilizes an autoregressive Transformer architecture~\cite{vaswani2017attention, chowdhery2023palm, brown2020language} and a single run step of LLM can be conceptualized as:
\begin{equation}
\begin{aligned}
   \text{output},\, KV^{\text{updated}} = \operatorname{LLM}(X,\, \text{indices},\, KV^{\text{past}})
    \label{eq1}
\end{aligned}
\end{equation}
In this process, input tokens $X$ are first transformed into an input matrix via an embedding table and then fed into the first layer. Subsequently, the input of each layer is the output of the previous layer. For each layer's input matrix, the input matrix is multiplied by three different weight matrices in the attention module to produce query, key, and value matrices, incorporating positional embeddings based on the input position indices. The LLM then multiplies query and key matrices to compute the attention scores, which represent the similarity between query and key, followed by scaling the scores and applying a mask before passing them through a softmax function to obtain the final attention weights. The generated key and value matrices are appended to the existing $KV^{\text{past}}$ to obtain the updated $KV^{\text{updated}}$, which accelerates future token generation. We refer to this intermediate result as the Key-Value Cache (KVCache) in this paper. Finally, the next token is sampled from the output of the model's last layer.

In real use cases, the inference process of LLM can be divided into two phases. 
In the prefill phase, the LLM processes all input tokens concurrently, using a position list $[1:L]$ (ranging from 1 to the length of inputs $L$) and an empty past KVCache. The output of this phase is the next token $X[L+1]$ and a KVCache of size $L  \times D  \times hd \times 2$. This process transforms $X[1:L]$ into $KV[1:L]$, which we refer to as \textbf{Full Attention (FA)}.
\begin{equation}
    X[L+1],\, KV[1\!:\!L] = \operatorname{LLM}(X[1\!:\!L],\, [1:L],\, \emptyset)
    \label{eq2}
\end{equation}

Following the prefill phase, the generation process transitions into the decoding phase. At time step $i$, the last generated token and all the past KVCache are fed to the LLM. The LLM outputs a new token and appends a matrix of size $1 \times D \times hd \times 2$. 
\begin{equation}
\begin{aligned}
    & X[L+i+1],\, KV[1\!:\!L\!+\!i] \\
    & = \operatorname{LLM}(X[L+i],\, L\!+\!i,\, KV[1\!:\!L\!+\!i\!-\!1]),\quad i = 1,2,\dots
    \label{eq3}
\end{aligned}
\end{equation}

As observed in the above equations, GPT-series LLM depends solely on previous tokens, a property known as causality. This causal property enables an effective method of reducing prefill cost called prefix KVCache: For any input tokens $X$ and a cached pair of previously processed input tokens $X^{\text{cached}}$ with its corresponding KVCache $KV^{\text{cached}}$, if there is a shared common prefix of length $s$ between $X$ and $X^{\text{cached}}$ (i.e., $X[1:s] = X^{\text{cached}}[1:s]$), we can directly reuse the first $s$ rows of $KV^{\text{cached}}$ to save computation:
\begin{equation}
\begin{aligned}
    & \operatorname{LLM}(X[1\!:\!L],\, [1:L],\, \emptyset) \\
    & = \operatorname{LLM}(X[s+1\!:\!L],\, [s+1:L],\, KV^{\text{cached}}[1\!:\!s])
    \label{eq4}
\end{aligned}
\end{equation}

This prefix KVCache reusing mechanism has been effectively applied in scenarios such as multi-turn conversations and shared agent system prompts. Examples include the local KVCache in vLLM~\cite{kwon_efficient_2023} and the distributed KVCache caching in Mooncake~\cite{qin2025mooncake}.

\subsection{KVCache Reuse in Retrieval Augmented Generation (RAG)} \label{sec:kvcrrag}
LLM typically generates responses based on pre-trained data, which may be outdated or insufficiently comprehensive to cover every domain. Patrick et al. \cite{lewis2020retrieval} introduced RAG, a method that retrieves external knowledge based on the user question, selecting the top-$n$ most relevant text chunks. One paradigm is to concatenate the system prompt ($S$), retrieved text ($C_{\text{top } i}$), and user question ($\mathcal{Q}$), and then feed the combined input into the model ($X=cat(S, \, C_{\text{top } 1}, \, ..., \, C_{\text{top } n}, \, \mathcal{Q})$), to standardize the output of model \cite{gao2023retrieval}.

In the context of RAG, related chunks are drawn from a shared knowledge base, creating a strong potential for reuse, especially for popular chunks. This suggests a significant opportunity for KVCache reuse. 

However, the strict requirement for an exact prefix match in previous schemes limits their applicability.
In RAG scenarios, since multiple chunks are retrieved together and can appear in \( n! \) possible orders for \( n \) chunks, the cache reuse ratio is low when relying on exact prefix matches. 
This issue persists even with advanced cache policies like RAGCache~\cite{Jin2024RAGCacheEK}.

\begin{figure}[t]
  \centering
    \includegraphics[trim=0 220 0 0, clip, width=0.7\linewidth]{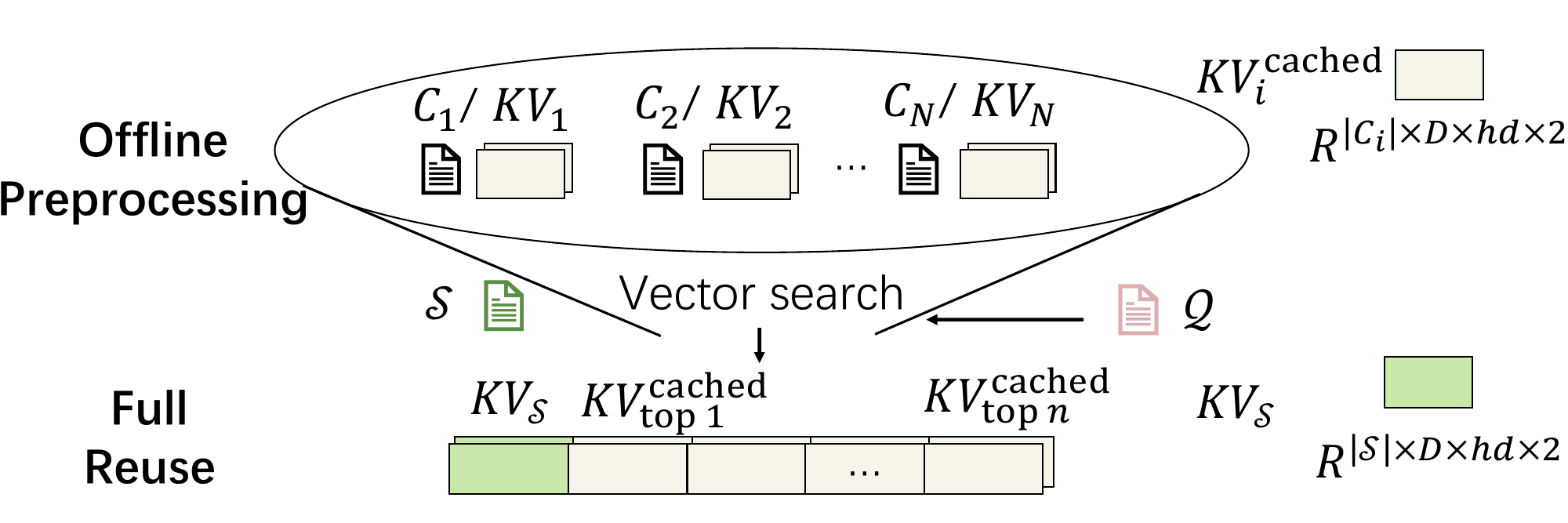}
    \vspace{-10pt}
    \caption{The preprocessing and reprocessing stage in Full Reuse and CacheBlend.}
    \vspace{-10pt}
  \label{fig:fullreuse_workload}
\end{figure}
\subsubsection{Full Reuse}
To address the low cache hit ratio of prefix cache in RAG scenarios, a straightforward solution is to treat each retrieved chunk as an isolated piece of knowledge that interacts only with the user question \( \mathcal{Q} \) and the system prompt \( \mathcal{S} \), as illustrated in the Figure \ref{fig:fullreuse_workload}. 

Specifically, the RAG execution can be divided into two stages: offline 
preprocessing and online reprocessing.

In the offline preprocessing stage, for a list of $N$ text chunks 
$[C_1, C_2, \dots, C_N]$ in the knowledge base, the system calculates 
and saves their corresponding KVCaches:
\[
[KV_1^{\text{cached}}, \, KV_2^{\text{cached}}, \, \dots, \, KV_N^{\text{cached}}],
\]
where each chunk is only attending to $\mathcal{S}$:
\begin{equation}
\begin{aligned}
     \_,\, KV &= \operatorname{LLM}(\operatorname{cat}(\mathcal{S},C_i),\, \left[1:\lvert \operatorname{cat}(\mathcal{S},C_i) \rvert\right],\, \emptyset), \\
    KV_{\mathcal{S}} &= KV\left[1:\lvert\mathcal{S}\rvert\right], \\
    KV_i^{\text{cached}} &= KV\left[\lvert\mathcal{S}\rvert+1:\lvert \operatorname{cat}(\mathcal{S},C_i) \rvert\right], \, i=1, \, \cdots, \, N
\end{aligned}
\label{eq5}
\end{equation}

Then, in the online processing stage, the cached KVCaches of all $n$ 
retrieved chunks are directly concatenated with each other to accelerate 
the subsequent prefill phase of the user question\footnote{During the stitching process of KVCache, each chunk of KVCache 
must be adjusted and embedded at the correct position. A complete example 
using RoPE~\cite{Su2021RoFormerET} is provided in Appendix \ref{appendix:a}}:
\[
[KV_{\text{top }1}^{\text{cached}}, KV_{\text{top }2}^{\text{cached}}, \dots, KV_{\text{top }n}^{\text{cached}}].
\]
\begin{equation}
\begin{aligned}
    & \_,\ KV^{\text{FA}} =\operatorname{LLM}(X,\ \left[1:L\right],\ \emptyset)  \\
    & \quad \Rightarrow \_,\ KV^{\text{FR}}= \operatorname{LLM}(\mathcal{Q},\ \left[|X|-|\mathcal{Q}|+1:|X|\right],\\
    & \quad \quad \quad \quad \quad \quad \quad \text{cat}(KV_{\mathcal{S}},\ KV_{\text{top }1}^{\text{cached}},\ \dots,\ KV_{\text{top }n}^{\text{cached}}))
\end{aligned}
\label{eq6}
\end{equation}

\begin{figure}[t]
  \centering
    \includegraphics[trim=0 65 0 35, clip, width=0.6\linewidth]{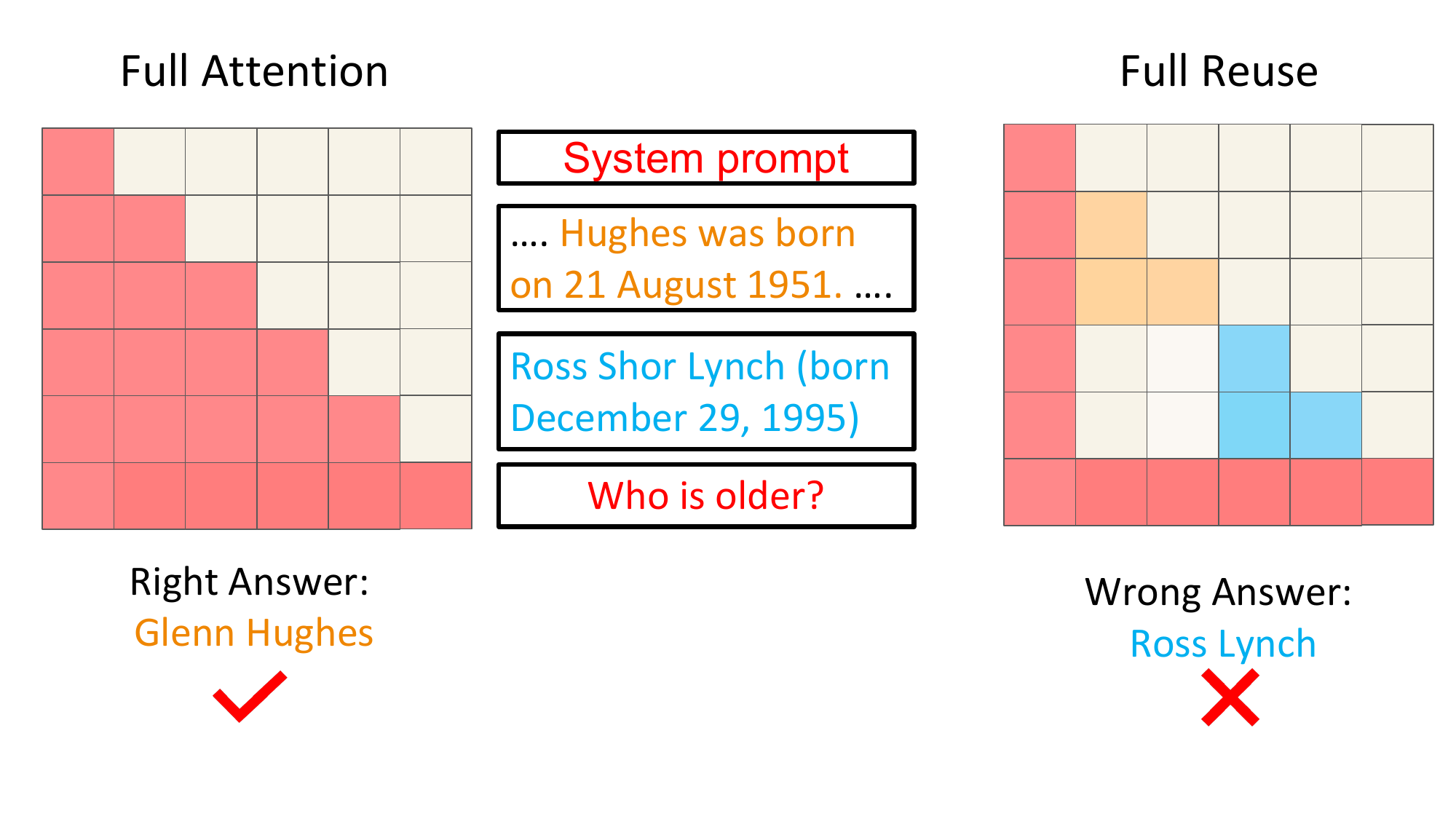}
    \vspace{-5pt}
    \caption{Full Attention is equivalent to a lower triangular mask. In contrast, Full Reuse is equivalent to Parallel Context Windows \cite{Ratner2022ParallelCW}, where each token only attends to earlier tokens within the same text chunk.}
    \vspace{-15pt}
  \label{fig:attentionMask}
\end{figure}

Note that in the above equation, we use an implication arrow (\(\Rightarrow\)) instead of an equality sign (\(=\)) because reusing \( KV_{\text{top }i}^{\text{cached}} \) {\bf is not an equivalent transformation} like the prefix cache.
It neglects all cross-attention between different retrieved chunks by applying a custom attention mask, as depicted in Figure \ref{fig:attentionMask}. In this approach, each retrieved chunk attends only to its own tokens and the system prompt \( \mathcal{S} \), while the user question \( \mathcal{Q} \) attends to all previous chunks. We refer to this strategy as {\bf Full Reuse (FR)}, which is utilized in systems like TurboRAG \cite{Lu2024TurboRAGAR} and PromptCache\cite{gim2024prompt}.

Full Reuse significantly increases the cache hit ratio because the KVCache of each chunk can be reused in any retrieval scenario, regardless of the order in which the chunks are retrieved.
While Full Reuse is highly efficient in reducing GPU costs, empirical results indicate a deterioration in generation quality during inference.
According to our experiments, the F1 score on Musique decreases up to 55\% after using Full Reuse.

The reason for the decline in generation quality with Full Reuse is the lack of cross-attention between text chunks, which can be reflected through KV deviation. The KV deviation \( \Delta_{KV} \) of size $L \times D \times 2$ on layer \( l \) is defined in fellowing equation, which measures the difference between \( KV^{\text{FA}}\)  and \( KV^{\text{\text{FR}}}\).
\begin{equation}
\begin{aligned}
    \Delta_{KV}[:, \, l] = \sum_{j=1}^{d} \left( KV^{\text{FA}}[:, \, l, \, j] - KV^{\text{\text{FR}}}[:, \, l, \, j] \right)^2
\end{aligned}
\label{eq7}
\end{equation}
KV deviation appears at the second layer and increases layer by layer. This change leads to incorrect answers, even resulting in nonsensical responses.

\subsubsection{CacheBlend}
To address the lack of cross-attention between chunks, \textbf{CacheBlend (CB)} proposes an innovative solution by introducing an online reprocessing stage that selectively reprocesses a subset of tokens. This approach significantly improves generation quality compared to the naive Full Reuse baseline, marking a crucial step toward practical chunk-level KVCache reuse. CacheBlend's core hypothesis is that, due to attention sparsity, fully compensating for cross-attention is unnecessary; recomputing only a critical subset is sufficient. Furthermore, CacheBlend leverages inter-layer consistency to identify this subset, pointing out that tokens with high KV deviation in one layer are likely to maintain high deviation in subsequent layers.

To implement this strategy, as shown in Figure \ref{fig:framework}, the online stage of CacheBlend is further partitioned into two sub-steps.

In the selection step, for the input $X = \operatorname{cat}(\mathcal{S}, \, C_{\text{top }1}, \, \cdots, \, C_{\text{top }n})$, CacheBlend applies both Full Attention and Full Reuse on the first two layers to obtain 
$KV^{\text{FA}}[:, \, 2]$ and $KV^{\text{FR}}[:, \, 2]$.
These two versions of KVCache are compared with each other to calculate  $\Delta_{KV}[:, \, 2, \, 1]$.  In the following equation, $\text{argTopk}(\cdot)$ returns the indices of the top-$k$ elements in a vector,
i.e., $\mathcal{C}_{\text{idx}}$ represents the indices of critical tokens in the input $X$.
\begin{equation}
\begin{aligned}
    \mathcal{C}_{\text{idx}}  = \operatorname{argTopk}(\Delta_{KV}\left[:, \, 2, \, 1\right])
\end{aligned}  
\end{equation}

In the second substep, CacheBlend performs sparse prefill. To do this, it constructs a new input prompt $\operatorname{cat}(X[\mathcal{C}_{\text{idx}}], \mathcal{Q})$ and a \textbf{special mask}, both of which are based on the positional indices of the critical tokens and the user question. This ensures that each token attends only to those preceding its position.
\begin{equation} 
\begin{aligned}
    \_, \, KV^{\text{CB}} = &   \operatorname{LLM}(\operatorname{cat}(X[\mathcal{C}_{\text{idx}}], \, \mathcal{Q}),\\
     & \text{cat}(\mathcal{C}_{\text{idx}}, \, [|X|+1:|X|+|\mathcal{Q}|]), \, KV^{\text{FR}}[1:|X|])\\
\end{aligned}         
\end{equation}

\begin{figure}[t]
  \centering
    \includegraphics[trim=20 310 10 0, clip, width=0.65\linewidth]{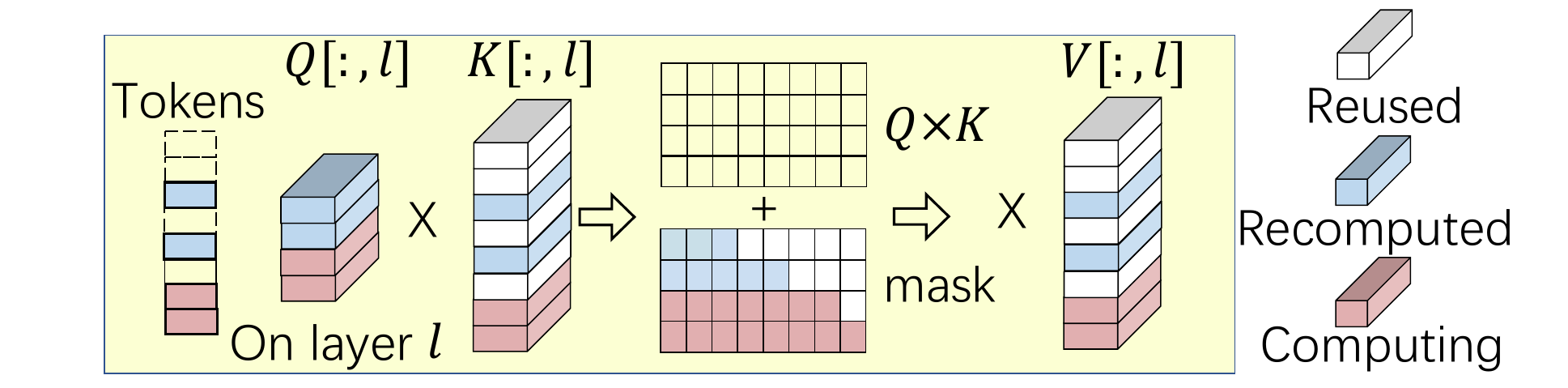}
    \vspace{-10pt}
    \caption{Cache Fusion performs self-attention over the critical tokens and the user query.}
    \vspace{-10pt}
  \label{fig:cacheBlend_workload}
\end{figure}
Figure \ref{fig:cacheBlend_workload} demonstrates a simple example of this process, where the original RAG input consists of the system prompt $X[1]$, two text chunks $X[2:3]$ and $X[4:6]$, and the user question $X[7:8]$. Supposing that the 3-rd and 5-th tokens are selected as the critical tokens,
CacheBlend will feed $[X[3], \, X[5], \, X[7], \, X[8]]$, position indices $[3, \, 5, \, 7, \, 8]$, and reused KVCache $KV[1:6]$ into the LLM for computation. 
To ensure proper casual attention, a mask of size $4\times8$ is constructed when calculating the attention scores. For $X[3]$, the corresponding mask$[1, \, 1:3]$ is set to 0, and mask$[1, \, 4:8]$ is set to $-\infty$, indicating that $X[3]$ can only attend to $X[1:3]$.

The sparse prefill of CacheBlend partially alleviates the issue of quality degradation, as shown in Figure \ref{fig:qwen2_series}. While it demonstrates some improvement in generation quality as the recomputation ratio increases, a significant gap remains between partial recomputation and Full Attention in some cases. For example, with 15\% recomputation using CacheBlend, Qwen2.5-7B-Instruct only recovers to the performance level of Qwen2.5-3B-Instruct on the TriviaQA and HotpotQA datasets.
We have identified two main reasons why CacheBlend performs poorly in certain scenarios: First, existing recomputation methods fail to efficiently restore quality without incurring significant latency. Secondly, the tokens selected by CacheBlend based on KV deviation are not the critical tokens that sparse attention focuses on. These findings highlight the need for a new approach. The central objective of our work is to maximize TTFT reduction while preserving generation quality.

\section{FusionRAG Framework}
\label{sec:FusionRAG_framework}
To achieve a better balance between quality and efficiency, we propose a novel inference framework called FusionRAG.
Our design is based on two key insights derived from the above observations.

\textbf{Insight 1:} \textit{Shift online reprocessing costs to the offline preprocessing stage for amortization.}

While increasing the ratio of critical tokens recomputed during reprocessing can improve generation quality, it also raises computational costs during online prefill, diminishing overall benefits. To keep the online recomputation ratio low, we focus on enhancing the preprocessing stage, where computations can be performed offline and their costs can be amortized through reuse across multiple inferences. This design philosophy aligns with the research approach in the compression domain, where compression itself requires time, but for datasets that are repeatedly used, the preprocessing costs can be amortized through subsequent reuse  ~\cite{10.1007/s00778-020-00636-3}.

While increasing the ratio of critical tokens recomputed during reprocessing can improve generation quality, it also raises computational costs during online prefill, diminishing overall benefits. To keep the online recomputation ratio low, we focus on enhancing the preprocessing stage, where computations can be performed offline and their costs can be amortized through reuse across multiple inferences. 

The crux here is that, since all retrieved chunks are semantically similar to the user question, they are also likely similar to each other. This relationship can be identified offline without knowledge of specific user queries. Thus, by grouping these similar chunks together, we enhance the needed cross-attention during the preprocessing stage by allowing them to attend to each other, thereby improving the quality of the cached representations.

\textbf{Insight 2:} \textit{The selection of critical tokens should be content-\hspace{0pt}dependent rather than position-dependent.} 

The critical token selection algorithm should not favor specific text chunks or tokens due to retrieval order or other inherent \textbf{biases}. Intuitively, the selection of critical tokens should be guided by the relevance between the user question and the retrieved chunk, leveraging the model’s own attention to identify the most informative parts of the text.

Based on these insights, FusionRAG operates in both of the two main stages. The modifications compared to CacheBlend are illustrated on the right side of Figure \ref{fig:framework}.

\subsection{Offline Stage: Similarity-Guided Preliminary Cross-attention Preprocessing}
\label{sec:offline_preprocessing}
To find semantically similar chunks, we utilize their similarity in the vector space. If vector \( A \) is highly similar to vector \( B \), and vector \( A \) is also similar to vector \( C \), then vectors \( B \) and \( C \) are likely similar as well. This transitive property allows us to group related text chunks based on their embeddings. With this in mind, we can use an embedding model to find the top-\( n \) similar documents for each text chunk, expecting that related chunks will likely be retrieved together (Co-occurrence) in response to user queries.

\begin{figure}[t]
    \centering
    
    \begin{subfigure}[b]{0.495\linewidth}
        \includegraphics[width=\linewidth]{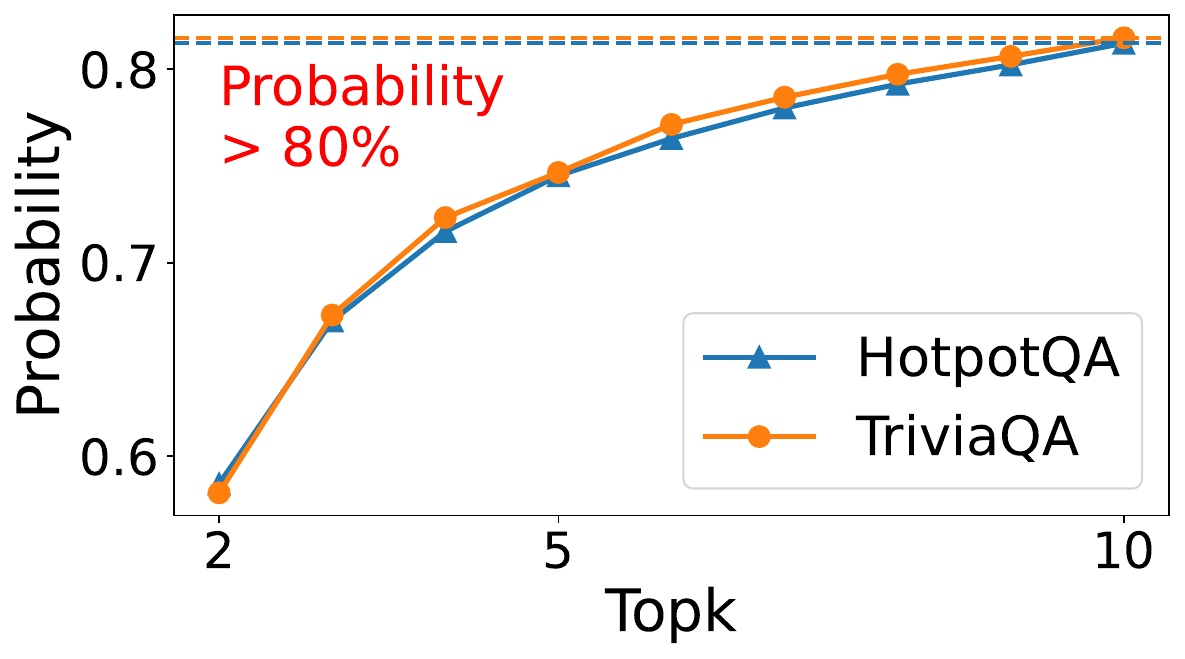}
        \caption{Performing RAG on text chunks to recall the probabilities of other text chunks under the same question.}
        \label{fig:offline_cdf}
    \end{subfigure}
    \hfill
    \begin{subfigure}[b]{0.495\linewidth}
        \includegraphics[trim=0 0 0 0, clip, width=\linewidth]{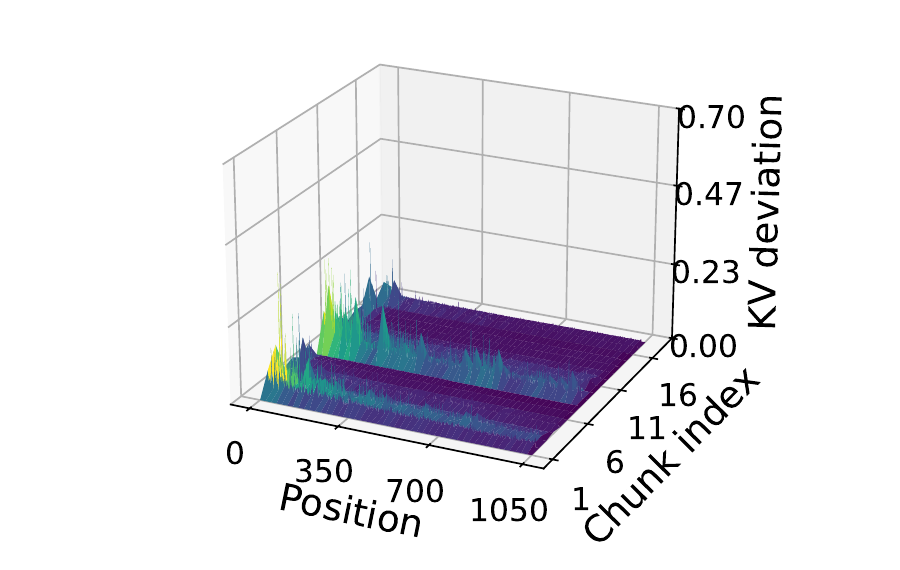}
        \caption{Similarity-Guided Preliminary Cross-attention Preprocessing can reduce KV deviation.}
        \label{fig:offline_deviation}
    \end{subfigure}
    \caption{The text chunks retrieved by RAG for the same question often exhibit a high similarity. The proposed offline preprocessing method effectively reduces KV deviation.}
    \vspace{-15pt}
    \label{fig:why_offline_good}
\end{figure}

To validate this assumption, we used BGEM3 \cite{Chen2024BGEMM} to retrieve the top-\( n \) relevant documents for each text chunk. Our findings, shown in Figure \ref{fig:offline_cdf}, indicate a high probability that RAG retrieves related text chunks for the same question. Even with only two text chunks recalled, the chance that RAG retrieves related blocks is over 50\%. With ten blocks recalled, this probability rises above 80\%. These results confirm that RAG tends to retrieve materials with high intrinsic relevance to each other for a given question.

Based on this characteristic, we enhanced the offline preprocessing stage as follows. We perform RAG on the text chunk $C_i$, retrieving the top-$n$ most relevant text chunks $[C_{i, \text{top } 1},C_{i, \text{top } 2}, \cdots, C_{i, \text{top } n}]$ along with their KVCache $[KV_{i, \text{top } 1}^{\text{cached}}, KV_{i, \text{top } 2}^{\text{cached}}, \cdots, KV_{i, \text{top } n}^{\text{cached}}]$.  Let $X=\operatorname{cat}(\mathcal{S},C_{i,\text{top } 1},\cdots,C_{i,\text{top } n})$. After concatenating these KVCache entries and inputting them into the model with positions set to $\left[|X|+1:|X|+|C_{i}| \right]$, we recompute the $C_i$'s KVCache $KV_{i}^{\text{cached}^\prime}$.
\begin{equation} 
\begin{aligned}
      \_, \, KV = & \operatorname{LLM}(C_i, \, \left[|X|+1:|X|+|C_{i}| \right], \\
     &  \text{cat}(KV_\mathcal{S}, \, KV_{i,\text{top } 1}^{\text{cached}}, \, \cdots, \, KV_{i,\text{top } n}^{\text{cached}})) \\
      KV_{i}^{\text{cached}^\prime} = & KV\left[|X|+1:|X|+|C_{i}| \right]
\end{aligned}
\label{eq10}
\end{equation}

We incorporate cross-attention in the preprocessing stage, thereby shifting part of the computational workload offline and bridging the gap between Full Reuse and Full Attention. This approach reduces the need for online reprocessing while enhancing generation quality. It effectively transfers some computational burden from the online stage to the offline stage, where costs can be amortized across multiple inferences. 

In Figure \ref{fig:offline_deviation}, we show the KV deviation of the model's second layer after similarity-guided preliminary cross-attention processing. Compared to Figure \ref{fig:cacheBlend_deviation_2_layer}, the offline preprocessing reduced the KV deviation of the second layer by 70\%, and this approach is not affected by position, as the KV deviation of all tokens decreases uniformly.

\noindent\textbf{Offline Preprocessing Cost Analysis:}
We quantify the computational overhead of the Similarity-Guided preprocessing stage to demonstrate its practicality. Using Qwen2.5-7B-Instruct with a batch size of 10, our measurements on the HotpotQA dataset show that the preprocessing stage processes text chunks at a rate of \textbf{0.218} seconds per chunk. This translates to a throughput of approximately 275 text chunks per minute, which is sufficient for most real-world RAG applications.

This overhead is practical and cost-effective for two reasons. First, the preprocessing is performed only once per document corpus and amortized across all subsequent queries, making the per-query overhead effectively zero. Second, the preprocessing stage is highly parallelizable and can be easily scaled across multiple GPUs or distributed computing resources, further reducing the wall-clock time in production deployments.

\noindent\textbf{Discussion:} While the discussion above focuses on embedding vector-based RAG\cite{Izacard2020LeveragingPR}, similarity-guided preliminary cross-attention processing can generalize to other paradigms. For example, in graph-based RAG\cite{Edge2024FromLT, Hu2024GRAGGR, Guo2024LightRAGSA}, each text chunk is represented as a vertex connected to related chunks by edges. These connections can help identify high co-occurrence probabilities, guiding preprocessing. A statistics-based approach could also analyze retrieval traces to anticipate co-occurrence. Integrating FusionRAG with these methods remains for future work.

\begin{figure}[t]
    \centering
    \begin{subfigure}[b]{0.24\linewidth}
        \includegraphics[width=\linewidth]{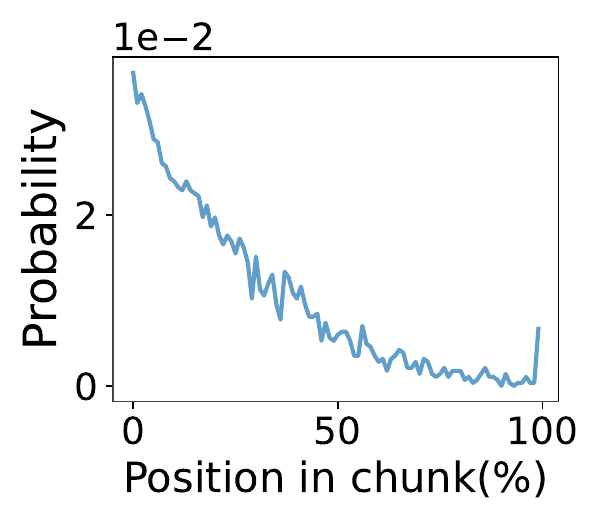}
        \caption{DCT within chunks in CacheBlend.}
    \end{subfigure}
    \hfill
    \begin{subfigure}[b]{0.24\linewidth}
        \includegraphics[width=\linewidth]{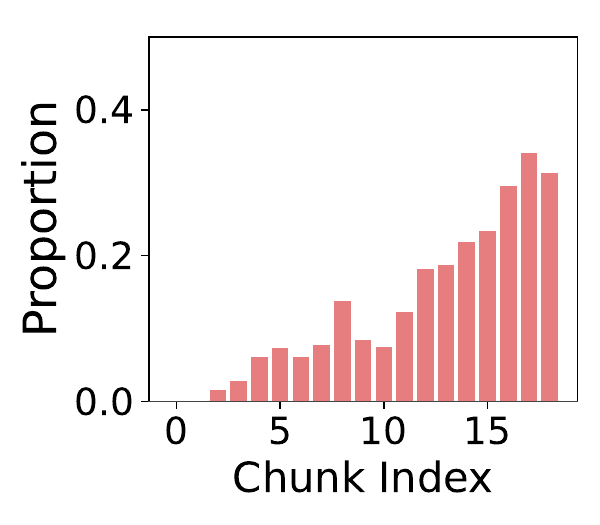}
        \caption{DCT across chunks in CacheBlend.}
    \end{subfigure}
    \hfill
    \begin{subfigure}[b]{0.24\linewidth}
        \includegraphics[width=\linewidth]{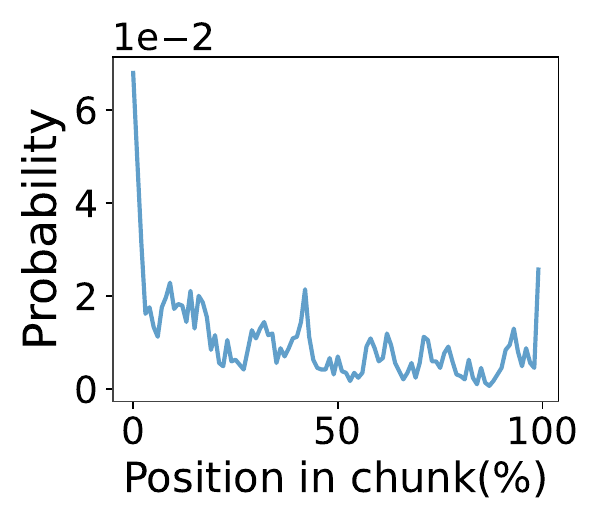}
        \caption{DCT within chunks in FusionRAG.}
    \end{subfigure}
    \hfill
    \begin{subfigure}[b]{0.24\linewidth}
        \includegraphics[width=\linewidth]{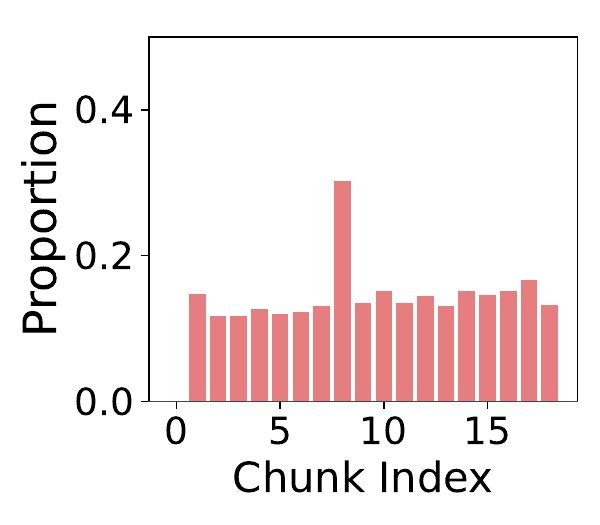}
        \caption{DCT across chunks in FusionRAG.}
    \end{subfigure}
    
    \vspace{-5pt}
    \caption{A visualization of the distribution of critical tokens (DCT) for a representative sample from Musique, illustrates the positions of 15\% critical tokens selected by CacheBlend and Query-Guided Selection within each chunk, as well as the selection ratio for each chunk.}
    \vspace{-10pt}
    \label{fig:3.1}
\end{figure}
\subsection{Online Stage: Query-Guided Selection}
\label{sec:online_reprocessing}
To select unbiased critical tokens, we propose a method based on attention weights, termed Query-Guided Selection. Prior work~\cite{Cai2024PyramidKVDK, jiang2024minference, li2024snapkv, liu2023scissorhands, zhang2023h2o} has shown that in the final layers of attention, certain columns of the attention weight matrix exhibit significantly higher values, forming vertical slash patterns. The corresponding tokens of these columns are particularly critical. Based on this observation, we propose the following Query-Guided Selection process: 

After the user submits a question, a prefill operation is performed on the question to generate the query matrix of the final layer. Then Query-Guided Selection computes attention weights by multiplying this query matrix and the key matrix of the final layer for each chunk. The attention weight is a matrix, where the number of rows and columns is the token count of the user question and the RAG text chunk, respectively. Subsequently, we sum each column of this matrix, resulting in a vector with a length equal to the token count of the retrieval text chunk, which is the critical score for our selection. After getting the critical scores of all chunks, the tokens with the top-$k$ highest critical scores are selected as the critical tokens.

The critical score of a token is independent of its position within the entire prompt or within its respective chunk. As shown in Figure \ref{fig:3.1}, we analyzed the selection of the top 15\% critical tokens by CacheBlend and Query-Guided Selection, examining their positional frequencies within chunks and their proportion of total tokens per chunk.
CacheBlend selects critical tokens in a manner that tends to favor the later chunks across the chunks, while within each chunk, it tends to focus on the head of the chunk. In contrast, Q-guided selection selects critical tokens in a relatively uniform manner across chunks. Although we observe that tokens near the beginning of each chunk tend to have higher critical scores, which aligns with the attention sink phenomenon reported in prior work \cite{Cai2024PyramidKVDK, Xiao2023EfficientSL}, the critical scores are not exclusively concentrated at the beginning.

\begin{figure}[b]
\vspace{-10pt}
  \centering
    \includegraphics[trim=0 0 0 0, clip, width=0.9\linewidth]{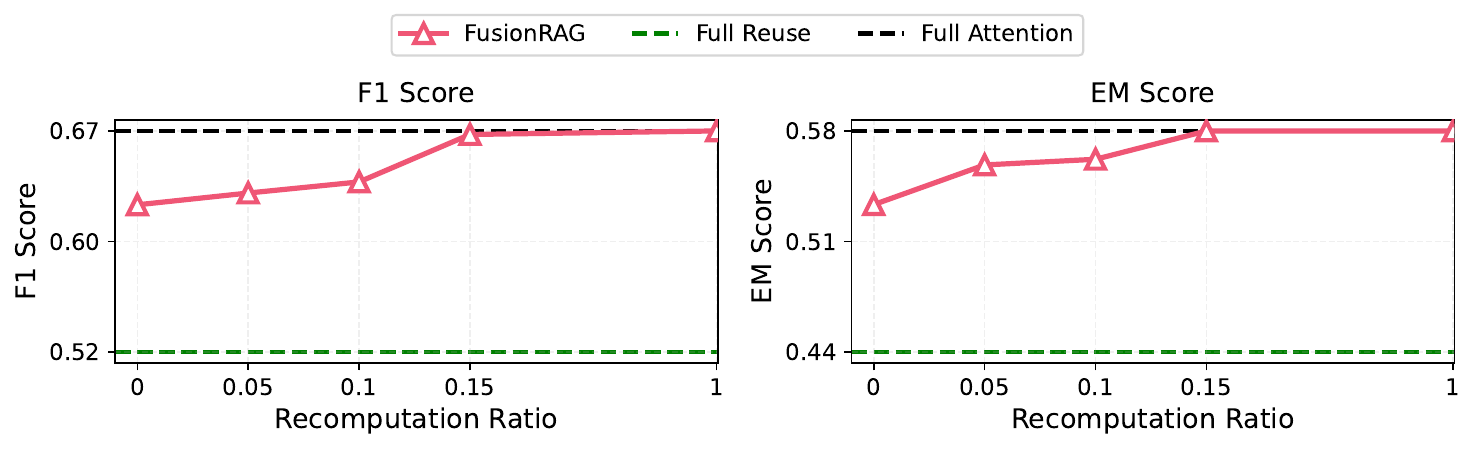}
    \vspace{-10pt}
    \caption{OpenPangu-1B with FusionRAG recovers Full Attention generation quality on TriviaQA.}
  \label{fig:openpangu}
\end{figure}

\textbf{Preliminary Validation.} To provide an early validation of the FusionRAG framework before comprehensive evaluation, we conduct a proof-of-concept experiment using OpenPangu-1B~\cite{Chen2025PanguEA} on the TriviaQA dataset. As illustrated in Figure \ref{fig:openpangu}, we assess FusionRAG's generation quality across different recomputation ratios using both F1 and EM metrics. The results demonstrate that on this compact 1B-parameter model, FusionRAG successfully recovers the generation quality of Full Attention under both F1 Score and EM metrics.

\input{chapter/system-design}

\section{Evaluation}
\label{experiment}

\subsection{Setup}
\label{sec:setup}

To verify the generality of the method, we validate our approach on the latest five open-source models: Mistral-7B-Instruct-v0.3 \cite{Jiang2023Mistral7}, Qwen2.5-7B-Instruct, Qwen2.5-14B-Instruct, Qwen3-32B and GLM4-32B~\cite{glm2024chatglm}, with Qwen3-32B and GLM4-32B executed using tensor parallelism (TP=4). The testing environment consist of 4 NVIDIA L20 GPUs (48GB VRAM), 1TB DRAM, and 7TB NVMe SSD.


\noindent\textbf{Baselines.} Across all experiments, we compare FusionRAG with the aforementioned existing schemes, i.e., Full Attention, Full Reuse, CacheBlend and Cache-Craft. 
Note that Full Reuse and Full Attention represent the two extremes with recomputation ratios of 0\% and 100\%, respectively, and position embeddings are adjusted following TurboRAG across all recomputation ratios.

\noindent\textbf{Benchmarks.} Four representative knowledge based QA datasets are used for mimicking real-world RAG scenarios.
\begin{itemize}
    \item \textbf{TriviaQA \cite{Joshi2017TriviaQAAL}:} This is a public dataset for the QA task, where text chunks are excerpted from Wikipedia and simple questions are posed for the LLM to answer. It is designated to test LLM’s 1-hop reasoning ability. The dataset we tested contains over 250 test cases.
    \item \textbf{HotpotQA\cite{Yang2018HotpotQAAD}:} To test the model's multi-hop reasoning capability, this dataset requires 2-hop reasoning across multiple Wikipedia pages to answer each question. It includes over 250 test cases.
    \item \textbf{Musique \cite{Trivedi2021MM}:} The Question in this dataset involves up to 4-hop reasoning.  We utilize the Musique dataset provided by LongBench \cite{Bai2023LongBenchAB}, which contains 200 questions.
    \item \textbf{2WikiMultiHopQA \cite{Ho2020ConstructingAM}:} This dataset consists of up to 5-hop questions. We use the 2WikiMultiHopQA dataset provided by LongBench, which contains 200 questions.
\end{itemize}

We evaluate FusionRAG across two dimensions:

\noindent\textbf{Quality Metrics (\textit{higher is better}):}
\begin{itemize}
    \item \textbf{Faithfulness}~\cite{ragas2024}: Evaluates whether the generated answer is faithful to the retrieved context without hallucinations.
    \item \textbf{Exact Match (EM)}~\cite{Rajpurkar2016SQuAD1Q}: Binary metric requiring exact string match with the ground-truth answer. This is the most stringent evaluation criterion.
    
    \item \textbf{F1 Score}~\cite{Rajpurkar2016SQuAD1Q}: Measures word-level overlap via precision and recall (computed using ROUGE-1~\cite{Bai2023LongBenchAB}). Compared to EM, F1 provides a more nuanced assessment by giving partial credit for overlapping tokens, typically resulting in higher scores.
    
    \item \textbf{Normalized F1 Score}: To facilitate cross-configuration comparison, we normalize F1 scores relative to two baselines:
    \begin{equation}
        \text{Normalized-F1} = \frac{\text{F1} - \text{F1}^{\text{FR}}}{\text{F1}^{\text{FA}} - \text{F1}^{\text{FR}}} \times 100\%
        \label{eq:normalized_f1}
    \end{equation}
    where $\text{F1}^{FA}$ and $\text{F1}^{FR}$ denote the F1 scores under Full Attention and Full Reuse, respectively. This normalization maps the F1 score to a percentage scale, indicating how much generation quality is preserved relative to the baseline methods. For Mistral-7B on TriviaQA, Full Attention achieves F1 = 0.852 (Normalized-F1 = 100\%), Full Reuse achieves F1 = 0.712 (Normalized-F1 = 0\%), and CacheBlend with 15\% recomputation achieves F1 = 0.781 (Normalized-F1 = 49.2\%).
\end{itemize}
\noindent\textbf{Efficiency Metrics (\textit{lower is better}):}
\begin{itemize}
    \item \textbf{TTFT}: End-to-end latency from query submission to first output token. Lower TTFT indicates better user experience.
\end{itemize}


\renewcommand{\arraystretch}{0.8}
\begin{table}[ht]
\centering
\vspace{-5pt}
\caption{Faithfulness on four datasets (15\% recomputation).}
\vspace{-5pt}
\begin{tabular}{l|c|c|c}
\toprule
\textbf{Dataset} & \textbf{CacheBlend} & \textbf{Cache-Craft} & \textbf{FusionRAG} \\
\midrule
\multicolumn{4}{c}{\textbf{Qwen3-32B}} \\
TriviaQA     & 0.7225 & 0.7271 & \textbf{0.7596} \\
HotpotQA     & 0.8334 & 0.8231 & \textbf{0.8597} \\
Musique      & 0.5180 & 0.5286 & \textbf{0.5457} \\
2WikiMQA     & 0.5923 & 0.5660 & \textbf{0.6410} \\
\midrule
\multicolumn{4}{c}{\textbf{GLM4-32B}} \\
TriviaQA     & 0.7312 & 0.7504 & \textbf{0.7624} \\
HotpotQA     & 0.8308 & 0.8555 & \textbf{0.8686} \\
Musique      & 0.3930 & 0.4954 & \textbf{0.5369} \\
2WikiMQA     & 0.5257 & 0.5768 & \textbf{0.5962} \\
\bottomrule
\end{tabular}
\label{tab:faithfulness_results}
\vspace{-10pt}
\end{table}

\begin{figure*}[t]
    \centering
        \includegraphics[width=0.9\textwidth]{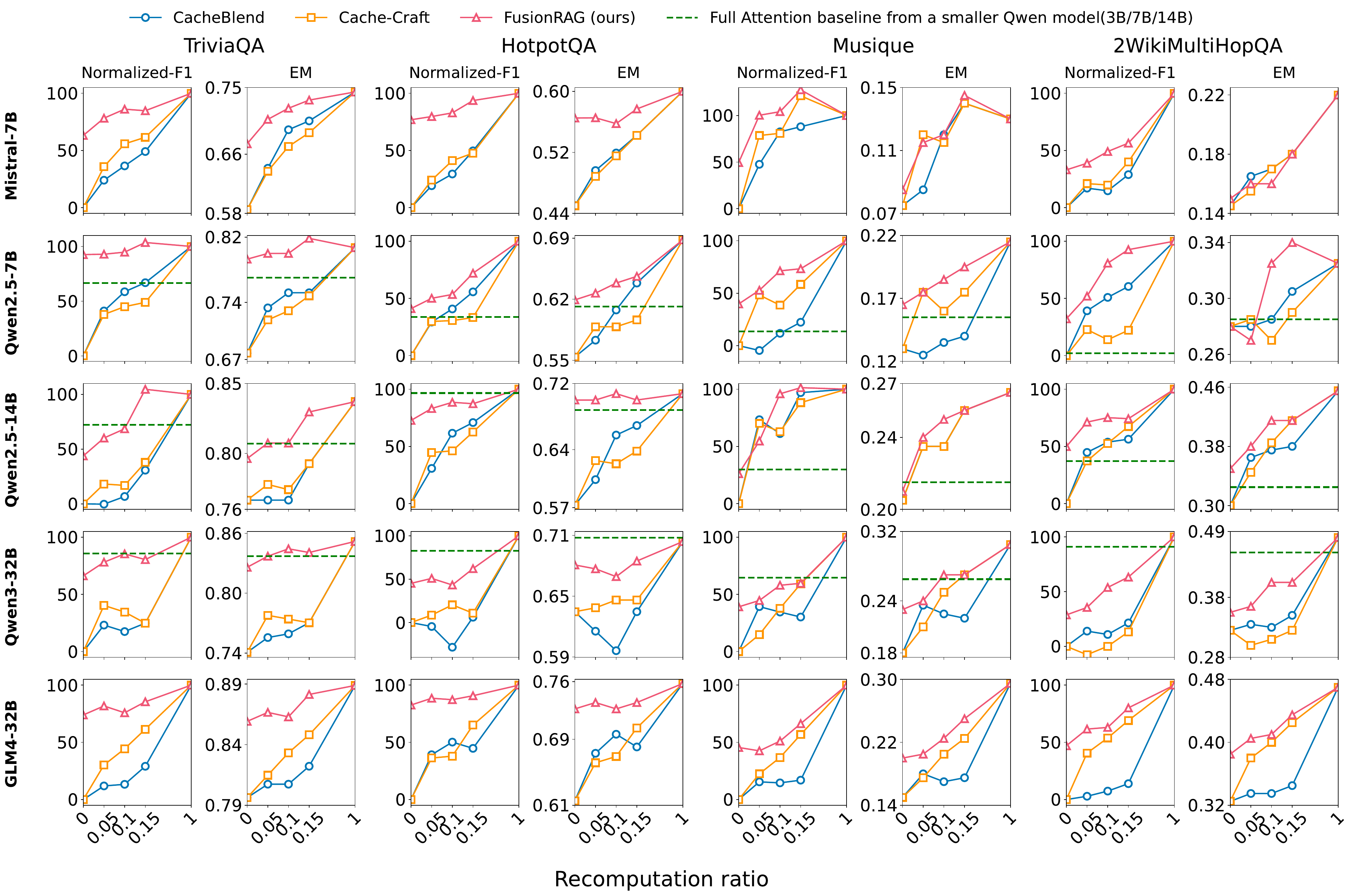}
    \vspace{-10pt}
    \caption{The Normalized F1 and EM scores of generated answers on single-hop and multi-hop datasets across five models.}
    \vspace{-10pt}
    \label{fig:generate_quality}
\end{figure*}

\subsection{Overall Improvement}
To demonstrate that FusionRAG achieves a better trade-off between generation quality and computation efficiency, we first compare all the systems at different recomputation ratios. 
Then we report the end-to-end speedup improvement brought by FusionRAG and further dissect this improvement by showing the efficiency of our Q-Sparse-Attn operator. 
Finally, we evaluate the throughput performance of FusionRAG in an end-to-end multi-question scenario.

\subsubsection{Better Quality/Efficiency Trade-off}
\label{sec:trade_off}

To comprehensively evaluate generation quality, we employ multiple metrics including Faithfulness, EM and F1. Our key findings are as follows:

\textbf{FusionRAG generates more faithful responses to the source documents.} Table \ref{tab:faithfulness_results} presents Faithfulness scores at 15\% recomputation ratio on two large-scale models (Qwen3-32B and GLM4-32B). FusionRAG consistently achieves the highest scores across all test cases, outperforming CacheBlend by up to 36.62\% and Cache-Craft by up to 13.25\%. This demonstrates that the Similarity-Guided preprocessing stage and the Query-Guided Selection reprocessing stage in FusionRAG do not induce model hallucinations but rather preserves stronger fidelity to the retrieved context.

\label{sec:experiment}
\begin{figure*}[t]
    \centering
    
    \begin{subfigure}[b]{0.445\linewidth}
        \includegraphics[trim=0 10 0 0, clip, width=\linewidth]{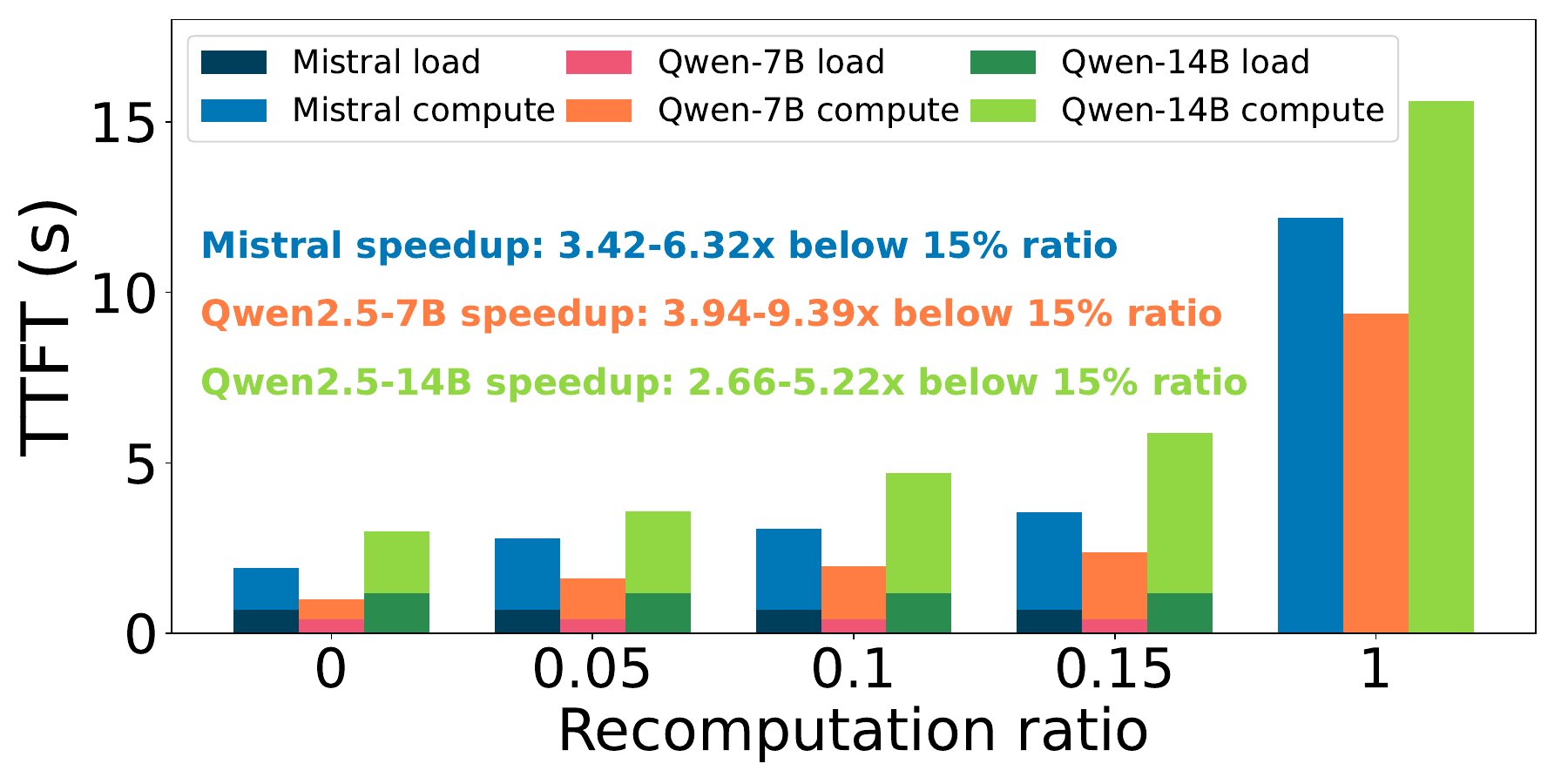}
        \caption{Visualization of end-to-end latency under different recomputation ratios.}
        \label{fig:end-to-end}
        \vspace{-10pt}
    \end{subfigure}
    \hfill
    \begin{subfigure}[b]{0.445\linewidth}
        \includegraphics[trim=0 10 0 0, clip, width=\linewidth]{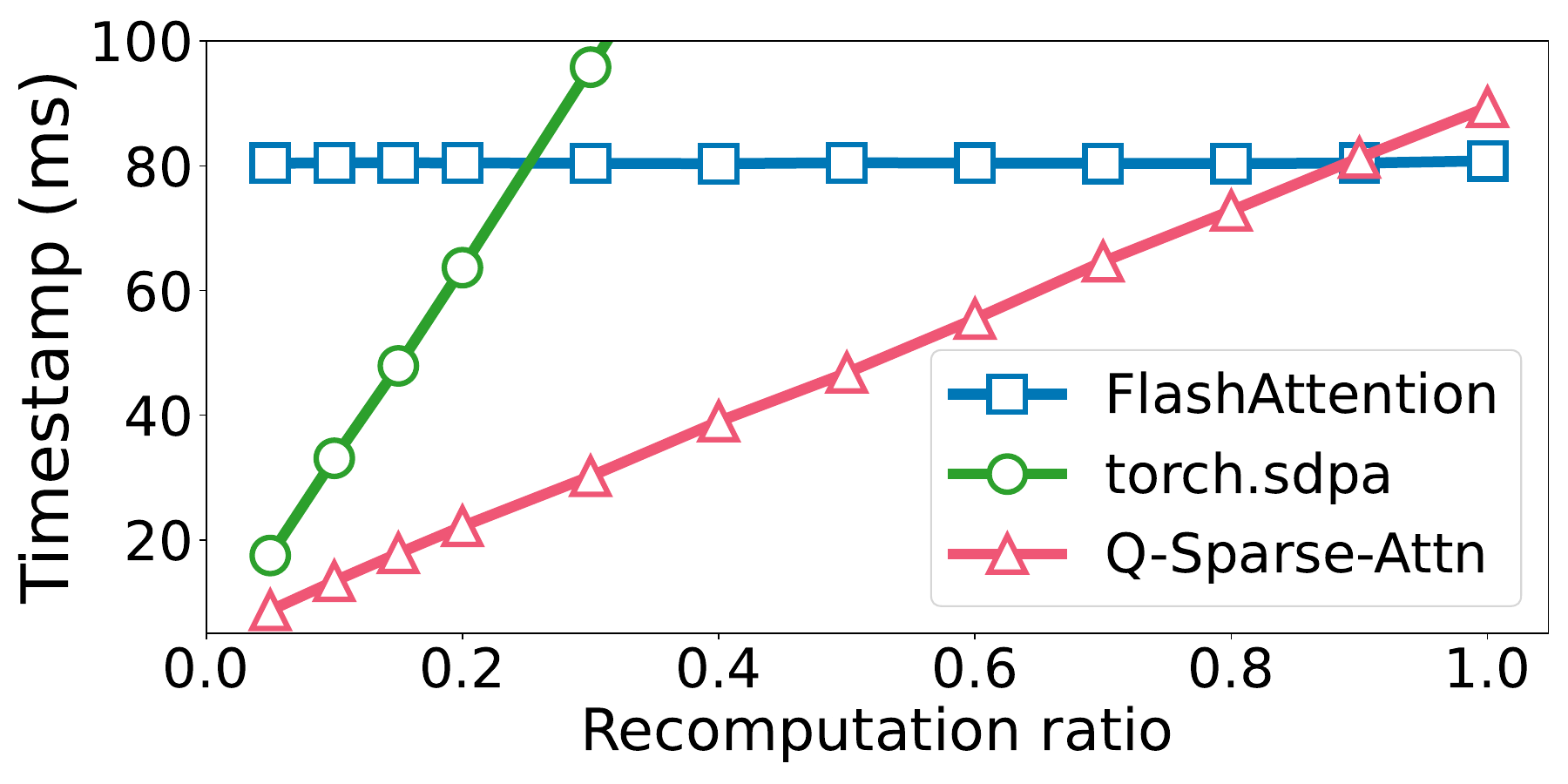}
        \caption{Performance comparison of Q-Sparse-Attn with existing operators.}
        \label{fig:Q-Sparse-Attn}
        \vspace{-10pt}
    \end{subfigure}
    \caption{Q-Sparse-Attn accelerates attention computation and reduces end-to-end latency.}
    \vspace{-10pt}
    \label{fig:why_offline_good}
\end{figure*}

\textbf{FusionRAG outperforms baseline methods in generation quality metrics.}
Figure \ref{fig:generate_quality} presents Normalized-F1 and EM scores across single-hop and multi-hop datasets. Although the scoring criteria for these two metrics differ (EM being strict, F1 more lenient), the quality trends they reflect are highly consistent across all experimental settings. FusionRAG demonstrates remarkable precision: at a 15\% recomputation ratio, even when measured by the strictest EM standard, 70\% of its scores (14/20) remain within 0.03 of the Full Attention baseline. This demonstrates that FusionRAG's generation quality effectively approaches that of Full Attention.

One notable outlier is the Qwen2.5-7B-Instruct model on 2WikiMultiHopQA, which exhibits a gap between its high F1 and low EM scores at 5\% recomputation. We attribute this divergence to a specific formatting issue: the model's responses are often semantically correct but incomplete. These answers receive partial credit (high F1) but fail the exact match criterion (low EM) because they are not the full, expected string.

\textbf{Breaking down results by query complexity shows that Similarity-Guided 
preprocessing delivers substantial gains on single-hop dataset, while 
FusionRAG's complete two-stage design maintains consistent robustness on 
multi-hop datasets.} For simple, single-hop dataset (TriviaQA), FusionRAG's offline preprocessing effectively mitigates quality degradation. This preprocessing is so effective that even with 0\% recomputation, FusionRAG already surpasses the performance of baseline methods operating at a 15\% recomputation ratio. FusionRAG consistently recovers at least 80\% of generation quality across five models at 15\% recomputation ratio. The challenge intensifies for complex, multi-hop queries, where baseline methods exhibit 
severe limitations at 15\% recomputation ratio. CacheBlend shows particularly 
poor performance on larger models (Qwen3-32B and GLM4-32B), achieving only 6\% 
recovery on Qwen3-32B HotpotQA. Cache-Craft also 
exhibits performance bottlenecks in certain scenarios, recovering only 11\% on 
Qwen3-32B HotpotQA. FusionRAG's two-stage design proves more robust: At 15\% recomputation ratio, FusionRAG maintains quality recovery ranging from 56\% to 100\% across all multi-hop datasets. 

Overall, 60\% of all tests (12/20) achieve over 80\% quality recovery at 15\% 
recomputation ratio, with FusionRAG consistently positioned closest to the ideal 
top-left corner in Figure \ref{fig:generate_quality}. FusionRAG achieves up to 
70\% higher Normalized-F1 scores than baseline methods at 15\% recomputation ratio. We observe minor local fluctuations in quality for all 
methods at certain recomputation ratios (e.g., Qwen2.5-14B useing CacheBlend on Musique shows lower F1 at 10\% than at 5\%), but the overall trend shows consistent improvement as recomputation increases.

\textbf{FusionRAG's generation quality can even surpass Full Attention in certain 
scenarios.} For instance, on Mistral-7B's Musique dataset, FusionRAG with 15\% recomputation achieves higher quality than Full Attention. This counter-intuitive result may be attributed to the ``overthinking'' phenomenon observed in prior work~\cite{Kaya2018HowTS, Pan2024EETuningAE}, where models produce correct answers in shallow layers but deeper computation may introduce errors. In our case, lower recomputation ratios maintain greater independence between text chunk KVCaches. As the recomputation ratio increases, additional cross-attention computation may introduce overthinking, potentially degrading performance in specific cases.



\subsubsection{Reduced TTFT} 
Figure \ref{fig:end-to-end} illustrates the TTFT of three models under different recomputation ratios, using a 27k token context, which is close to the typical context length. The KVCache loading time refers to the time taken to transfer data from the host memory to the GPU memory. 
Compared to Full Attention, at a 0\% recomputation ratio, the model's TTFT can be reduced by 5.22-9.39$\times$, while at a 15\% recomputation ratio, the model's TTFT can be reduced by 2.66-3.94$\times$. 

The improvement in TTFT is mainly attributed both to KVCache Reuse and advancements in the attention operator. Figure \ref{fig:Q-Sparse-Attn} compares the performance of Q-Sparse-Attn with SOTA attention operators in recomputing critical tokens under the 32k token context. FlashAttention does not support custom mask or positional indices for the query matrix, and thus can only perform casual mask attention,  followed by filtering critical tokens for the KVCache. Consequently, the computation time in Figure \ref{fig:Q-Sparse-Attn} remains constant regardless of the recomputation ratio. In contrast, Q-Sparse-Attn achieves a 4.2$\times$ speedup when recomputing 15\% of the tokens compared to FlashAttention. We also tested the SDPA operator, which enables recomputing critical tokens via custom mask and defaults to the xformer \cite{xFormers2022} backend in our test environment. Compared to SDPA, Q-Sparse-Attn provides a 2.69$\times$ speedup when recomputing 15\% of tokens.

When using Full Attention (with a recomputation ratio of 1), Q-Sparse-Attn is slightly slower than FlashAttention due to the additional memory access for the Q indices.

\begin{figure*}[t]
    \centering
        \includegraphics[width=0.95\textwidth]{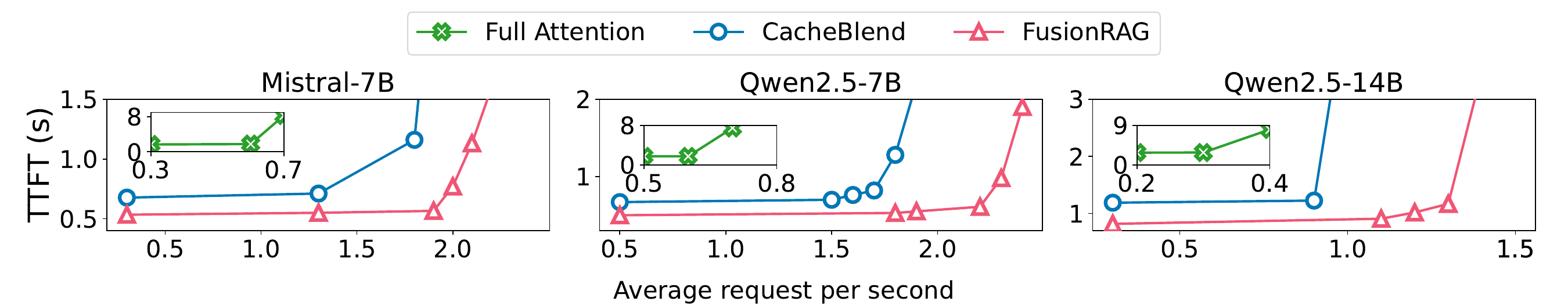}
    \vspace{-10pt}
    \caption{FusionRAG achieves higher throughput while maintaining generation quality closer to that of Full Attention.}
    \vspace{-10pt}
    \label{fig11}
\end{figure*}
\subsubsection{Throughput Improvement}\label{sec:throughput}

To demonstrate the benefit of FusionRAG in continuous batching with high concurrency, we simulated a multi-question concurrent scenario by setting up a thread to send the aforementioned test requests to the LLM at varying request rates. As shown in Figure \ref{fig11}, both FusionRAG and CacheBlend employ a 15\% recomputation ratio. CacheBlend follows its open-source 
implementation, using the SDPA operator for sparse prefill computation and without employing the Asynchronous KVCache Scheduler proposed 
in this work. Compared to all the baselines across different models, FusionRAG demonstrates 1.2-4.3$\times$ throughput.

\subsection{Ablation studies}

To quantify the contribution of each component in FusionRAG, we conduct comprehensive ablation experiments that address three key questions: {\em 1)} Does FusionRAG's Similarity-Guided preprocessing generalize to other token selection methods? {\em 2)} What is the optimal configuration for the hyperparameters (top-$n$ and recomputation ratio)? {\em 3)} What is the contribution of each optimization component?

\begin{figure}[t]
        \includegraphics[width=0.7\linewidth]{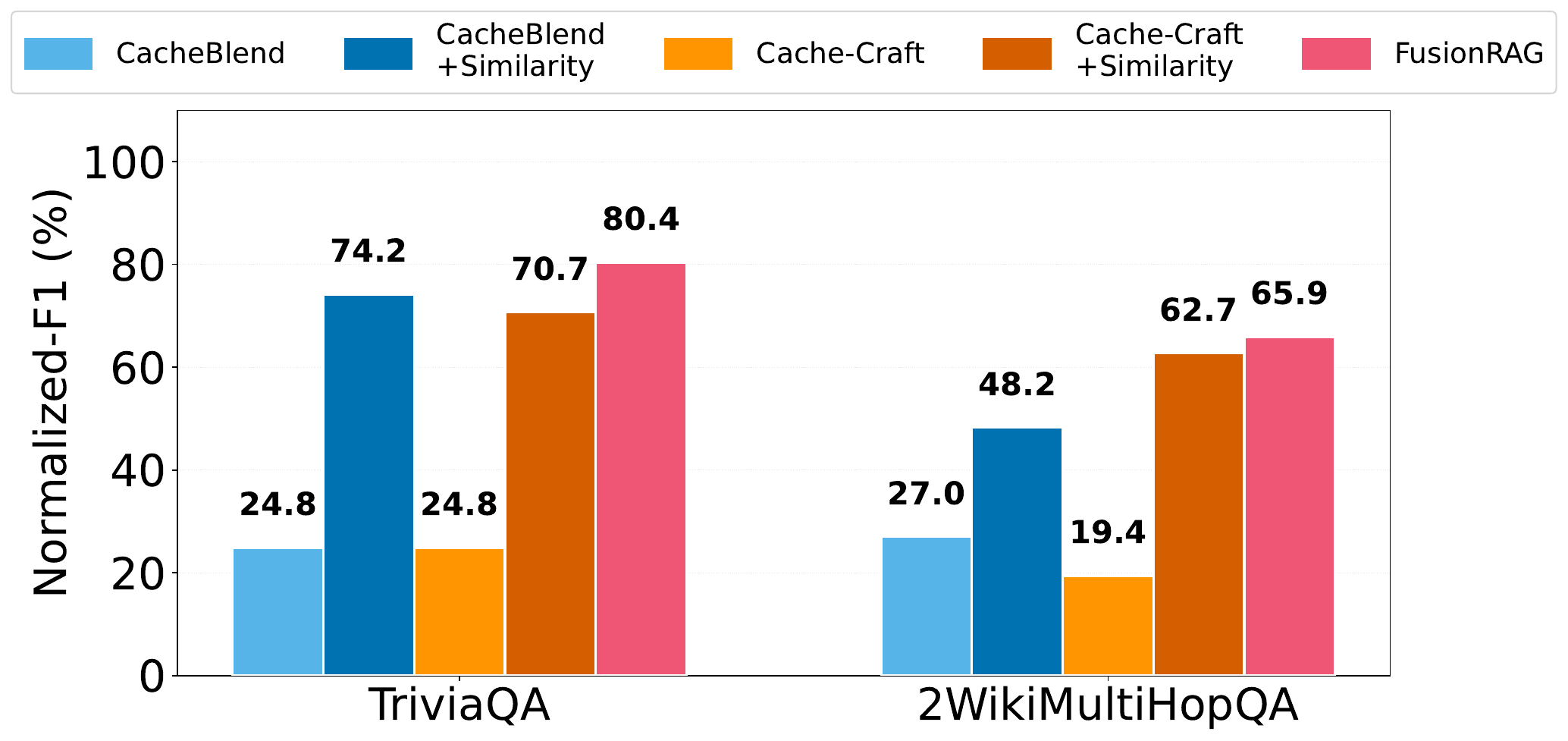}
        \vspace{-10pt}
    \caption{The Generation Quality of Combinations of Different Selection Methods and Similarity-Guilded preprocessing stage.}
    \label{fig:selection_compare}
    \vspace{-10pt}
\end{figure}

\subsubsection{Selection Strategy Comparison}
\label{sec:selection_comparison}

We select two scenarios where CacheBlend and Cache-Craft perform poorly at 15\% recomputation ratio and apply our offline preprocessing stage to them. As shown in Figure \ref{fig:selection_compare}, applying Similarity-Guided preprocessing stage to baseline methods yields substantial generation quality improvements of 21.2\%--49.4\% for CacheBlend and 43.3\%--45.9\% for Cache-Craft across TriviaQA and 2WikiMultiHopQA datasets, demonstrating that our offline preprocessing stage effectively reduces KV deviation and benefits various selection strategies. Furthermore, even when enhanced with preprocessing, FusionRAG with Query-Guided Selection consistently achieves the highest generation quality, outperforming preprocessed baselines by 3.2\%--17.7\%. This result confirms that the combination of Similarity-Guided preprocessing and Query-Guided Selection within FusionRAG yields the most effective approach.

\begin{figure}[t]
        \includegraphics[width=\linewidth]{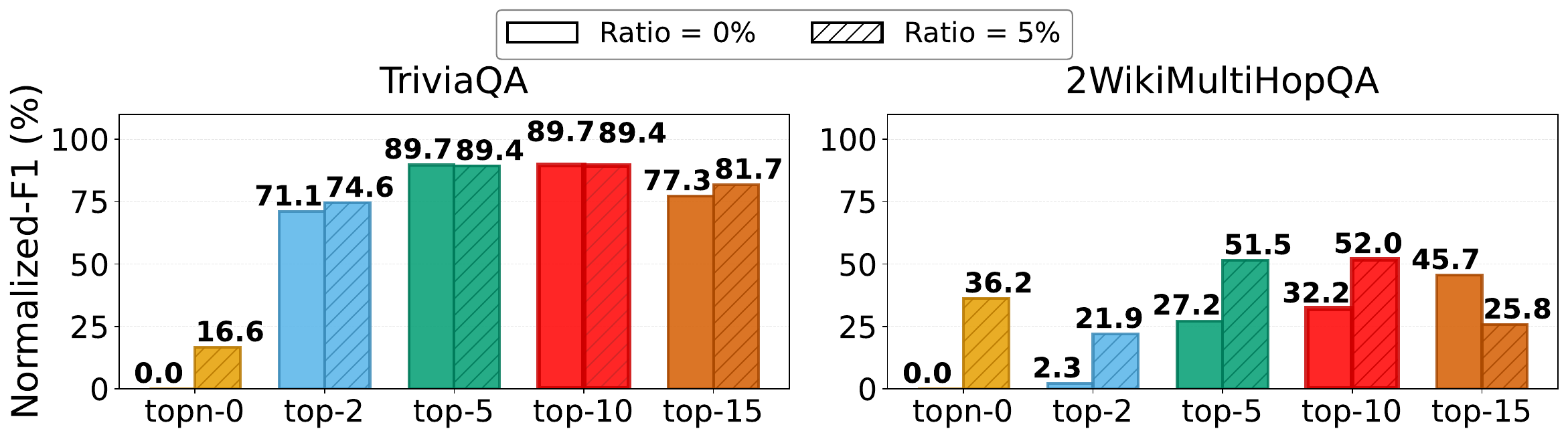}
        \vspace{-20pt}
    \caption{Effect of top-$n$ hyperparameter on generation quality.}
    \label{fig:topn_breakdown}
    \vspace{-10pt}
\end{figure}

\subsubsection{Hyperparameter Sensitivity}
\label{sec:hyperparameter_sensitivity}
The FusionRAG framework involves two key hyperparameters: \textit{\textbf{${\text{top-}n}$}} similar text chunks in the offline stage and the \textit{\textbf{recomputation ratio}} in the online stage.

$\bm{\text{\textbf{top}-}n}$ \textbf{Analysis.} We recommend $\text{top-}n=10$ as the default configuration. As shown in Figure \ref{fig:topn_breakdown}, we compare the performance of different $\text{top-}n$ values on Qwen2.5-7B across two datasets. On the single-hop dataset TriviaQA, $\text{top-}10$ achieves the best performance, whereas $\text{top-}15$ already shows a decline in generation quality. On the multi-hop dataset 2WikiMultiHopQA, although $\text{top-}15$ performs best at a 0\% recomputation ratio, its performance likewise drops when the ratio is increased to 5\%. Both phenomena indicate that further increasing the top-n value (e.g., to 15) may introduce noise and lead to a deterioration in generation quality. Therefore, we conclude that $\text{top-}n=10$ is the optimal configuration.

\textbf{Recomputation Ratio Analysis.} For the online recomputation ratio, we adopt 15\% as the default configuration. This choice is motivated by two considerations: First, it aligns with the baseline setting in CacheBlend to ensure fair comparison. Second, as shown in Figure~\ref{fig:generate_quality}, 15\% recomputation recovers over 80\% of the quality loss in 60\% of the test scenarios (12/20), while further increasing the ratio incurs linear growth in computational overhead with diminishing marginal quality gains. Therefore, 15\% represents the optimal trade-off between efficiency and quality.

\subsubsection{Impact of Individual Components}
\label{sec:each_component_impact}

\begin{figure}[t]
        \includegraphics[width=0.75\linewidth]{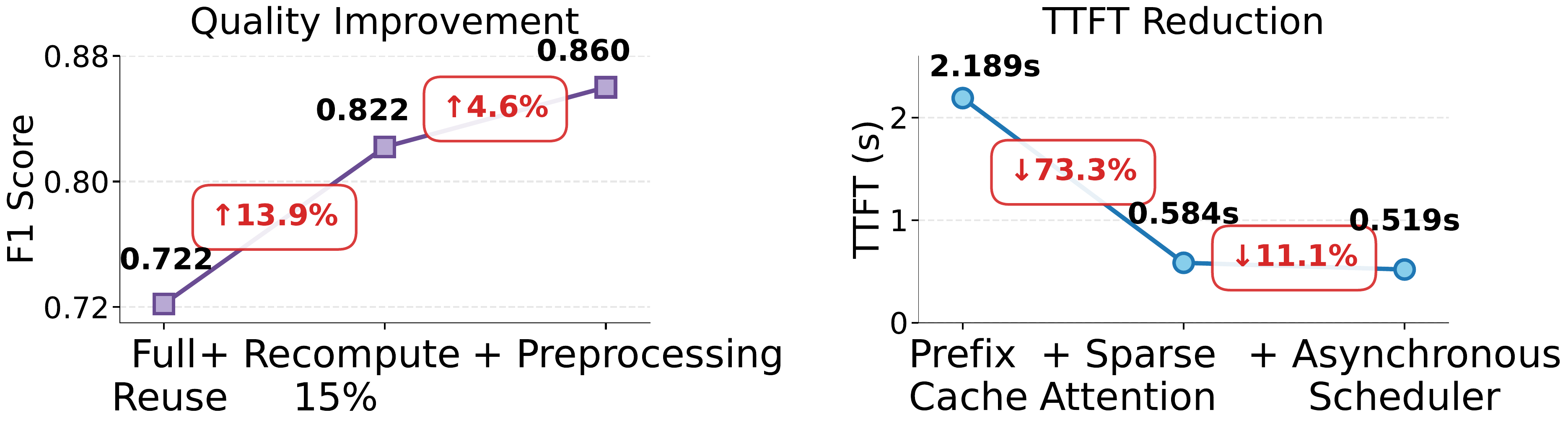}
        \vspace{-10pt}
    \caption{The impact of each component on the TTFT and generation quality of the system.}
    \label{fig:ablation_study}
    \vspace{-10pt}
\end{figure}

We conduct comprehensive ablation experiments to evaluate each core component across three dimensions: generation quality, TTFT, and storage efficiency. As shown in Figure \ref{fig:ablation_study}, components are categorized by optimization targets: {\em 1)} Similarity-Guided preprocessing and Query-Guided selection for quality, and {\em 2)} Q-Sparse-Attn and Asynchronous KVCache Scheduler for TTFT. The Alternative Path mechanism, targeting storage efficiency, is analyzed separately.

\textbf{Generation Quality.} On TriviaQA with Qwen2.5-7B-Instruct, we measure recovery between Full Reuse (0.722 F1) and Full Attention (0.869 F1). Query-Guided Selection alone achieves \textbf{72\%} quality recovery (0.822 F1). Adding Similarity-Guided preprocessing reaches \textbf{99\%} recovery, nearly matching Full Attention performance.

\textbf{TTFT Acceleration.} With batch size 10, Q-Sparse-Attn at 15\% recomputation reduces TTFT by \textbf{73.3\%} versus Full Attention baseline (2.189s). The Asynchronous KVCache Scheduler provides an \textbf{additional 11\%} reduction, confirming both components are essential for prefill efficiency.

\textbf{Storage Efficiency.} We simulate 1,000 queries where 60\% of chunks already exist in the knowledge base, 30\% are shared among users but newly uploaded, and 10\% are completely unique. Compared to a baseline using standard prefix cache, FusionRAG reduces total KVCache storage by 71.0\%. While the baseline recomputes the same chunk's KVCache 4,036 times across different prefix contexts, FusionRAG eliminates 82.3\% of these redundant computations through Alternative Path matching. The mechanism achieves a 77.9\% hit rate, confirming its effectiveness in identifying and reusing existing KVCache across diverse query patterns.

These ablation results demonstrate that each component provides a distinct and significant contribution to FusionRAG's overall performance, validating our system design choices.

\begin{figure}[t]
        \includegraphics[width=0.6\linewidth]{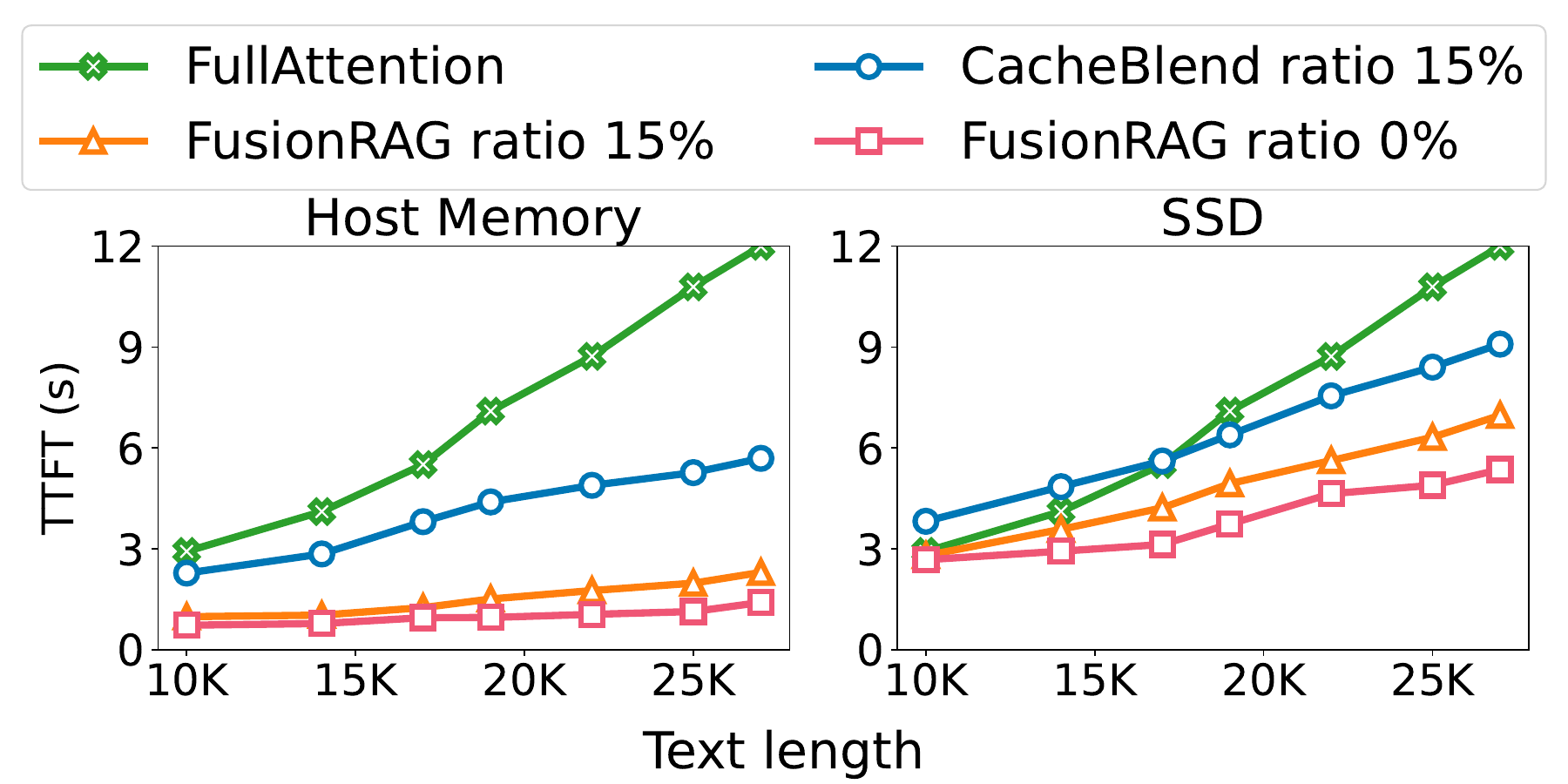}
        \vspace{-10pt}
    \caption{FusionRAG outperforms baselines with different text lengths when using host memory and SSD.}
    \vspace{-10pt}
    \label{fig12}
\end{figure}

\subsection{System Analysis}
For a better understanding of FusionRAG, we further analyze how varying configurations impact performance.

\textbf{Varying sequence lengths:} Figure \ref{fig12} (left) shows the computation time comparison for FusionRAG with text lengths ranging from 10K to 27K using Mistral-7B, considering both 0\% recomputation and 15\% recomputation scenarios. As the text length increases, the computation time for FusionRAG at a recomputation rate of 0\% remains consistently low($\le1.4$s). With a recomputation rate of 15\%, FusionRAG achieves a 2.3-3$\times$ speedup compared to CacheBlend in terms of TTFT. Additionally, FusionRAG reduces TTFT by 2.9-5.4$\times$ compared to Full Attention.

\textbf{Varying KVCache storage device:} To evaluate the impact of different storage devices on FusionRAG, we stored the KVCache of each method on an SSD and tested the computation time for text lengths ranging from 10K to 27K. The results are shown in the Figure \ref{fig12} (right). When the text length is below 17K, CacheBlend with 15\% recomputation offers no performance improvement, whereas FusionRAG still demonstrates a significant performance advantage under the same conditions. 

\section{Related Work}
\label{sec:relatedwork}
Beyond the KVCache reuse strategies in \S\ref{sec:introduction} and \S\ref{sec:back}, many other methods aim to reduce the storage and computation cost of the KVCache.

RAGCache \cite{Jin2024RAGCacheEK} is orthogonal to our work as it increases cache hit ratio through prefix tree construction but still employs standard prefix caching mechanisms. PromptCache \cite{gim2024prompt} targets system messages and prompt template scenarios However, in RAG scenarios, prompt templates constitute a relatively small proportion of the entire context. 

Another orthogonal direction addresses KVCache reuse through model fine-tuning. Methods such as Block Attention \cite{Ma2024BlockAttentionFE}, TurboRAG \cite{Lu2024TurboRAGAR}, and KVLink \cite{Yang2025KVLinkAL} adapt LLMs to local attention patterns. However, fine-tuning demands substantially more compute than inference alone and requires careful dataset curation to balance chunk configurations and task diversity. These limitations motivate our training-free approach.

A separate line of research exploits the observation that not all layers contribute equally to generation quality. These budget allocation mechanisms leverage this heterogeneity to distribute memory based on component importance, balancing resource usage against accuracy: Layer-wise methods \cite{Cai2024PyramidKVDK, Yang2024PyramidInferPK, Ye2025InfiniteRA, Zhou2024DynamicKVTA, zhang2024simlayerkvsimpleframeworklayerlevel} assign compression ratios at the layer level, while head-wise approaches \cite{Feng2024AdaKVOK, Feng2025IdentifyCK, Zhang2024UnifyingKC, Tang2024RazorAttentionEK, Fu2024NotAH, Xiao2024DuoAttentionEL} offer finer control across individual attention heads.

Other work focuses on identifying redundancy within or across layers \cite{Wang2024ModelTY, wan2024look, Wan2024D2ODD, Nawrot2024DynamicMC, Wang2024LoMALC, Yang2024KVSharerEI, Liu2024MiniCacheKC, zhang2024unifying}, reducing repeated patterns in cached attention. By compressing or reusing data that appears multiple times, these methods shrink memory footprint without compromising quality.

\section{Limitations}
\label{sec:limitations}
We acknowledge the following limitations of our method. 
FusionRAG's primary limitation occurs in scenarios requiring frequent, large-scale corpus updates. While preprocessing adds only 0.218s per chunk (\S\ref{sec:offline_preprocessing}), systems with continuous high-volume updates (e.g., live news feeds ingesting thousands of documents hourly) face non-negligible cumulative overhead conflicting with freshness requirements. FusionRAG excels when documents are queried repeatedly, amortizing preprocessing costs, but may underperform in "write-once, read-once" scenarios. It is best suited for stable, frequently-queried corpora (e.g., enterprise knowledge bases).

\section{Conclusion}

In this paper, we introduce FusionRAG, a novel framework that extends prefix-based KVCache reuse to RAG tasks. Building upon existing two-stage RAG cache reuse frameworks, we introduce modifications to both the offline preprocessing and online recomputation stages. Existing methods are unable to maintain generation quality while accelerating the prefill. To compensate for the degradation in quality, we propose a two-stage approach. In the offline stage, we embed information from similar text chunks into each chunk based on their similarity. During the online stage, we simulate the model’s attention distribution over the user query to identify critical tokens and selectively recompute their KVCache.
Moreover, existing methods face compatibility issues with prefix cache and batch decoding. To address this, we first introduce an Alternative Path mechanism to enable compatibility with the prefix cache. Second, we design an asynchronous KVCache and scheduler to mitigate GPU idle time caused by I/O blocking during KVCache reads. Finally, we redesign and optimize the sparse attention operator to support FusionRAG’s reprocessing stage and enable efficient batch decoding. Our experiments show that, at similar recomputation ratios, FusionRAG significantly improves generation quality, achieving EM scores comparable to Full Attention standard across four QA benchmarks. By overlapping I/O blocking and using a sparse recomputation operator, FusionRAG also enhances computational efficiency, reducing TTFT by 2.66--9.39$\times$ with a recomputation ratio of less than 15\%. Overall, it provides an optimal balance between generation quality and computational overhead in RAG scenarios.

\section{Acknowledgments}
We thank the anonymous reviewers and our shepherd for their valuable 
comments and suggestions. The authors affiliated with Tsinghua University 
are all in the Department of Computer Science and Technology, Beijing 
National Research Center for Information Science and Technology (BNRist), 
Tsinghua University, China. This work is supported by National Key Research 
\& Development Program of China (2024YFB4505600), Natural Science Foundation 
of China (92467102) and Tsinghua University Initiative Scientific Research 
Program, Young Elite Scientists Sponsorship Program by CAST (2022QNRC001), 
Beijing Natural Science Foundation (L252014), and Ministry of Education of 
China Scientific Research Innovation Capability Support Project for Young 
Faculty (SRICSPYF-ZY2025022).

\input{chapter/appendix}
\bibliographystyle{ACM-Reference-Format}
\bibliography{AI}


\end{document}

%% file: chapter/system-design.tex
\section{FusionRAG System Implementation}

In this section, we discuss three important aspects of how to implement the FusionRAG framework in an LLM serving system, including:
{\em 1)} How to eliminate redundant KVCache storage for identical chunks across different prefix contexts while maintaining compatibility with existing prefix caching;
{\em 2)} How to avoid GPU waiting issues due to the I/O blocking for KVCache reads;  
{\em 3)} How to implement an efficient sparse attention kernel suitable for FusionRAG's recomputation stage.

Existing systems, like vLLM~\cite{kwon_efficient_2023}, use a prefix cache to facilitate KVCache reuse. This mechanism leverages a hashmap that maps prefixes to page indices: when multiple queries share the same prefix, the computed hash for that prefix points to the same KVCache page, thereby avoiding redundant computations.

\begin{figure}[t]
    \centering
    \includegraphics[width=0.45\linewidth]{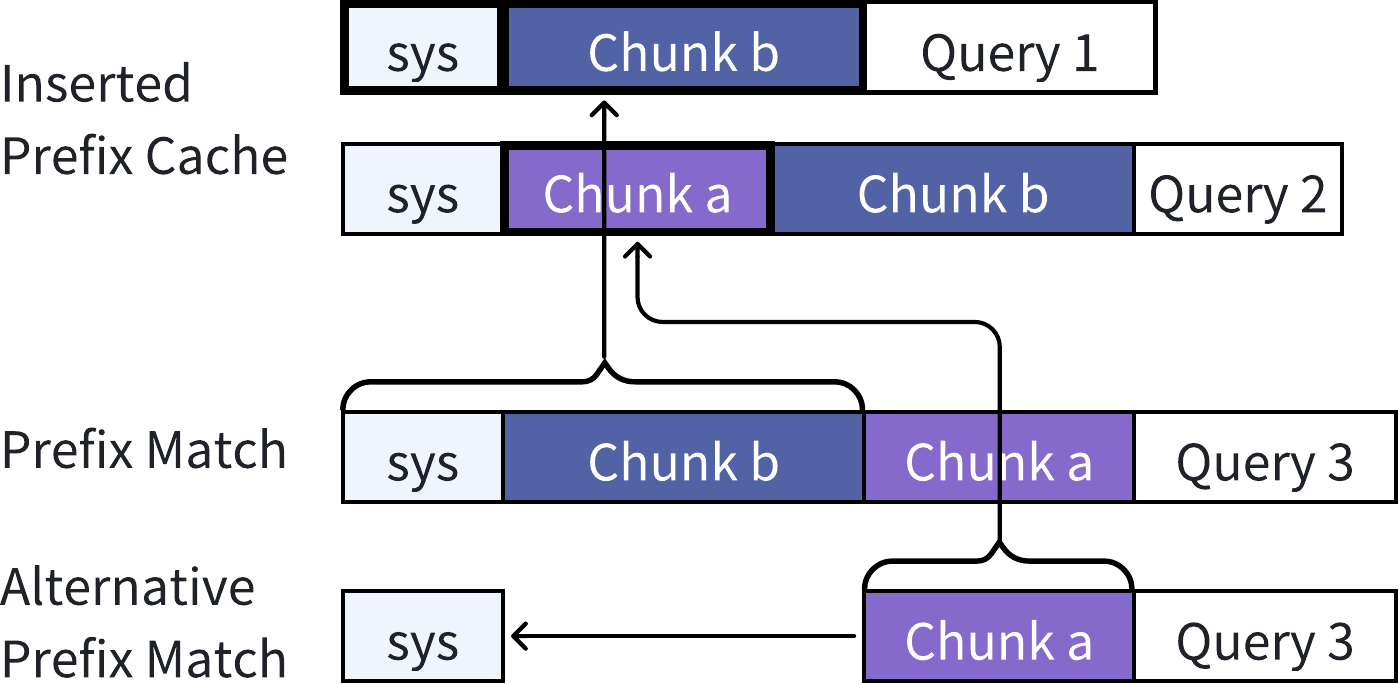}
    \vspace{-5pt}
    \caption{Query 3 uses alternative path to match Chunk $a$.}
    \vspace{-10pt}
    \label{fig:alt-path}
\end{figure}

However, in RAG scenarios, the prefix cache has a critical limitation: \textbf{identical text chunks in different prefix contexts cannot be recognized and reused.} As shown in Figure~\ref{fig:alt-path}, Query1 and Query2 have cached the KVCache for Chunk $b$ and Chunk $ab$ respectively. When Query3 also contains Chunk $b$ and Chunk $a$, the prefix matching logic only identifies Chunk $b$ but fails to match Chunk $a$—even though Chunk $a$'s KVCache was already computed and cached when processing Query2.

This limitation leads to severe storage waste in two-stage RAG systems: the offline stage only preprocesses chunks in the knowledge base, while the online stage may encounter new chunks outside the knowledge base. If these new chunks are frequently accessed by multiple queries (i.e., hot chunks), the standard prefix cache will create multiple KVCache copies for the same chunk across different prefix contexts, resulting in significant storage overhead and redundant computations.

To facilitate a higher cache reusing ratio while reducing storage overhead, our goal is to design a method that can locate previously cached chunks (e.g., Chunk $a$ in Query3) without modifying the existing prefix cache structure, enabling drop-in compatibility with current systems. To achieve this, we introduce the {\bf Alternative Path} mechanism. The core idea is: if a chunk has been cached before, its KVCache should be reused regardless of the preceding context, rather than being recomputed. The mechanism employs a progressive backtracking strategy: starting from the longest prefix, if matching fails, it iteratively removes earlier chunks to create shorter candidate paths until a match is found. For example, in Query3, the prefix ``system prompt + Chunk $b$ + Chunk $a$'' fails to match. FusionRAG then backtracks, skipping Chunk $b$ to form the shorter alternative path ``system prompt + Chunk $a$'', which successfully computes the correct hash and matches the KVCache stored from Query2.

Alternative Path ensures each unique chunk maintains only one KVCache copy, eliminating duplication while remaining fully compatible with existing prefix cache implementations, enabling seamless integration into current systems.

\subsection{Prefix Cache with Alternative Path}
\label{sec:alternative_path}

\subsection{Asynchronous KVCache Scheduler}
\label{sec:async_kvcache_scheduler}
In RAG scenarios, accessing KVCache from slower storage devices (disk or host memory) significantly increases overhead and blocks inference execution. This stems from the frequent migration of retrieved text chunks across the storage hierarchy. Due to limited device memory capacity, KVCache generated during the offline preprocessing stage must be stored in disk or host memory, then dynamically loaded to the GPU when needed. To reduce the loading overhead across the DISK-CPU-GPU hierarchy, FusionRAG employs an Asynchronous KVCache Scheduler that overlaps KVCache loading, request scheduling, and inference execution.
FusionRAG restructures the system into three asynchronous parallel components:

(1) \textbf{Scheduler}: The scheduler is responsible for request scheduling and KVCache management across a three-tier storage hierarchy (DISK-CPU-GPU). Upon receiving a user request, it first performs KVCache matching: if the required KVCache resides on disk or host memory, the scheduler dispatches an asynchronous loading request to the KVCache Loader; once the KVCache is ready in GPU memory, the request is enqueued into the execution queue. Throughout this process, KVCache is managed based on access frequency---when GPU memory is insufficient for new allocations, cold KVCache entries are evicted to host memory, and similarly from CPU to disk, following a heat-based tiering policy.

(2) \textbf{KVCache Loader}: The loader continuously loads KVCache from disk or host memory to the GPU in the background.

(3) \textbf{Inference Engine}: The inference engine continuously processes ready requests from the execution queue. It employs continuous batching~\cite{yu2022orca} to dynamically merge multiple requests and improve GPU utilization.

\begin{figure}[t]
    \centering
    \begin{subfigure}[b]{0.48\linewidth}
        \includegraphics[width=\linewidth]{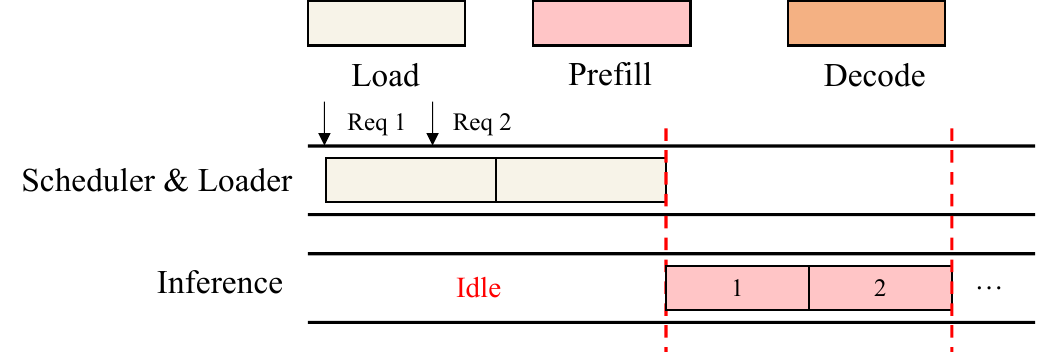}
        \caption{w/o Asynchronous Scheduler.}
        \label{fig:sync}
    \end{subfigure}
    \hfill
    \begin{subfigure}[b]{0.48\linewidth}
        \includegraphics[width=\linewidth]{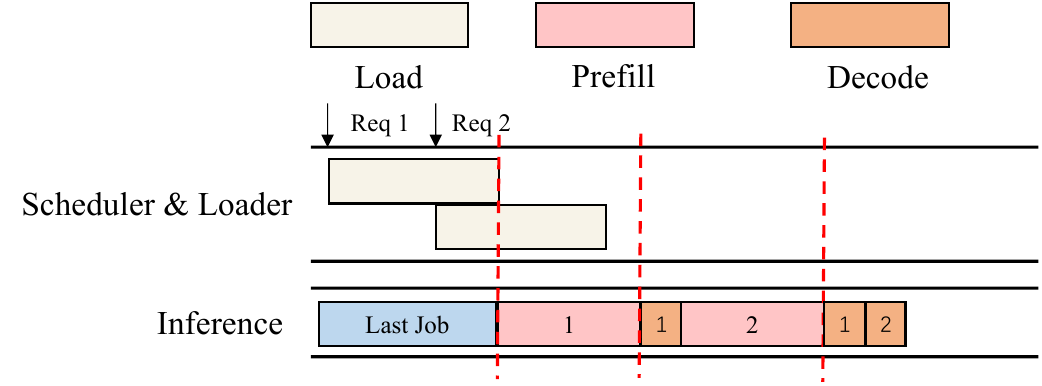}
        \caption{Asynchronous Scheduler.}
        \label{fig:async}
    \end{subfigure}
    
    \vspace{-8pt}
    \caption{Asynchronous KVCache scheduler overlaps request scheduling, KVCache loading, and inference.}
    \label{fig:async_workflow}
    \vspace{-10pt}
\end{figure}

Figure~\ref{fig:async_workflow} shows how asynchronous execution hides I/O latency. Figure~\ref{fig:sync} shows a synchronous baseline where the GPU remains idle during KVCache loading. When Req1 and Req2 arrive, the system sequentially loads their KVCache from disk. The inference engine can only begin prefill after both loading operations complete, resulting in significant GPU idle time.

In contrast, Figure~\ref{fig:async} demonstrates the asynchronous design of FusionRAG. When Req1 arrives, the Scheduler dispatches a loading task to the KVCache Loader while the Inference Engine continues processing ongoing jobs. Once the KVCache of Req1 is ready in GPU memory, the engine immediately begins its prefill stage. Crucially, while the KVCache of Req2 is being loaded, the Inference Engine concurrently executes the prefill stage of Req1. When the KVCache of Req2 becomes available, it is batched with Req1 using continuous batching to maximize GPU utilization. This pipeline design eliminates GPU idle time by overlapping I/O operations with computation, ensuring sustained GPU utilization throughout execution.

\noindent\textbf{Discussion on Asynchronous Loading Strategies:} 
Prior works~\cite{10.1145/3725273, Gao2024CostEfficientLL} propose layer-wise KVCache loading strategies that asynch\-ronously preload layer $l+1$ from disk while the GPU executes layer $l$, achieving fine-grained I/O-computation overlap. To address bandwidth-constrained scenarios, these approaches also incorporate prefetching mechanisms to load KVCache data further in advance. FusionRAG adopts request-level asynchronous loading for two key reasons:

\textbf{(1) Storage constraints in RAG scenarios:} RAG systems often manage large-scale knowledge bases in bandwidth-constrained storage systems. To store massive KVCache volumes for extensive document collections, production systems need to deploy remote distributed storage infrastructures, where network I/O bandwidth becomes the limiting factor. When $T_{load} \gg T_{prefill}$ (i.e., I/O bandwidth becomes the bottleneck), layer-wise and request-level overlap converge in behavior. Taking Qwen2.5-7B-Instruct (28 layers) as an example: processing 27K tokens on L20 requires 0.07s per layer, while each layer's KVCache occupies 0.05GB. With network bandwidth of 100 MB/s (typical in shared storage deployments), the system must prefetch 25 out of 28 layers—effectively loading the entire chunk's KVCache before inference begins, which is equivalent to request-level overlap. In such bandwidth-constrained scenarios, request-level loading reduces I/O fragmentation and minimizes network transfer overhead.

\textbf{(2) System design alignment:} Request-level loading naturally aligns with the RAG retrieval workflow, where the operational granularity is document chunks rather than model layers. This design simplifies KVCache management and integrates seamlessly with existing RAG pipelines.

We acknowledge that layer-wise loading can achieve finer-grained overlap when storage bandwidth is sufficient to sustain per-layer computation. The trade-off between these strategies depends on the system's I/O characteristics and workload patterns.

\begin{figure}[t]
    \centering
    \includegraphics[width=0.4\linewidth]{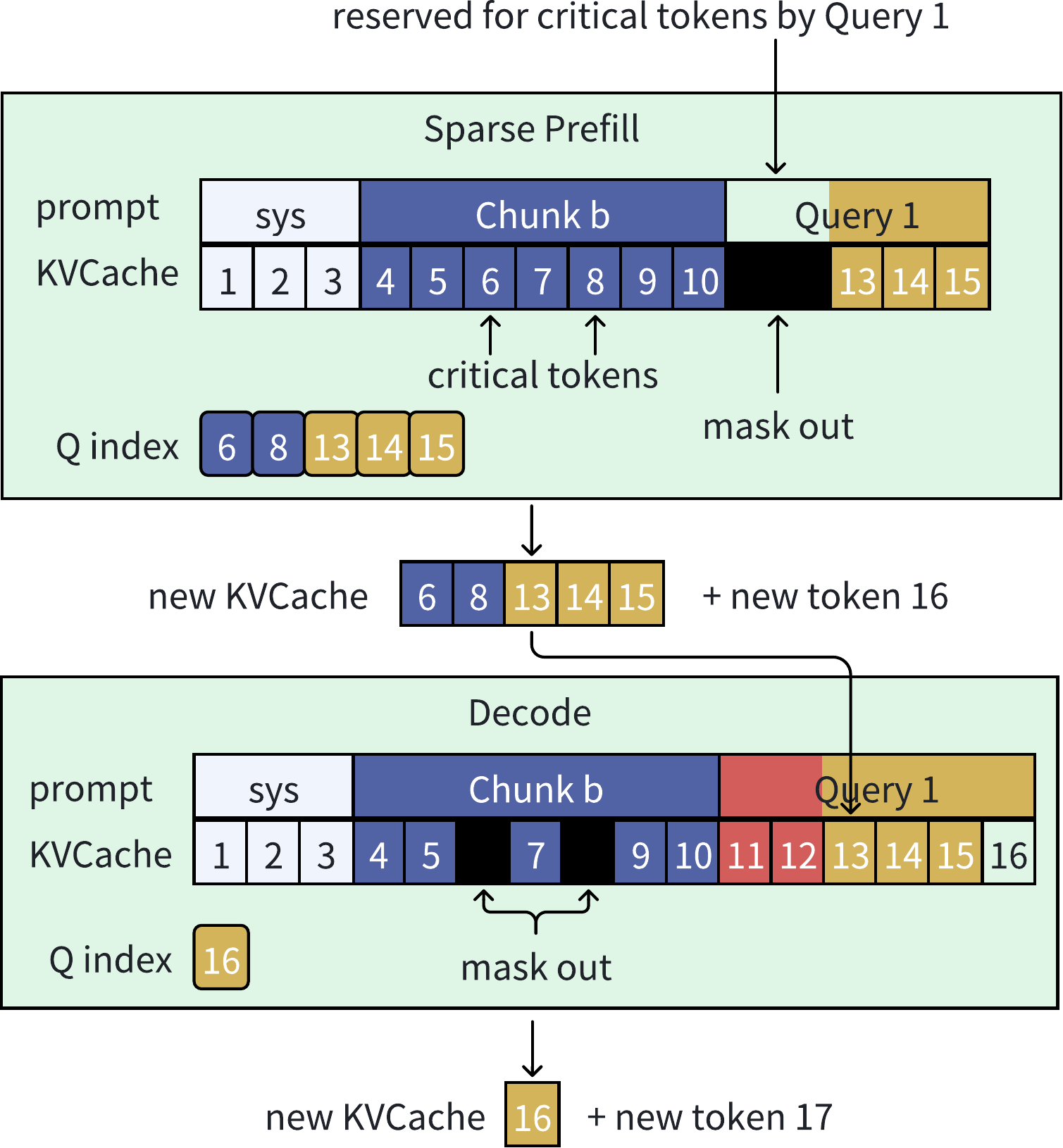}
    \vspace{-8pt}
    \caption{Q index is used to support sparse attention.}
    \label{fig:q-index}
    \vspace{-15pt}
\end{figure}

\subsection{Sparse Attention Optimization}
The efficiency of FusionRAG relies heavily on token-level sparse attention during both prefill and decoding. This presents a challenge in batch processing: multiple queries can share the same original KVCache for a chunk, but each query requires a different set of critical tokens to be recomputed. Modifying the original (shared) KVCache in-place would corrupt it for other queries in the batch. Therefore, to preserve the shared cache, we write the recomputed KVCache (for the critical tokens) to an exclusive page for each query. A mask is then used to exclude the original KVCache positions of these critical tokens within the shared chunk. This ensures the attention mechanism ignores these stale entries and instead computes attention using the newly computed KVCache from the exclusive page, while still accessing all non-critical tokens from the original shared cache.

However, existing attention operators are not optimized for this kind of sparse attention. Many of these operators either do not support sparse attention or lack performance optimizations. For example, FlashAttention~\cite{dao2022flashattention} does not support attention mask and requires contiguous KVCache. 
FlexAttention~\cite{flexattention} requires explicit mask creation (along with some auxiliary variables). In our experiments with a 32K-length text and 15\% recomputation, we observed that mask creation in FlexAttention requires 16GB of memory and takes 2s, which are too expensive.

To address this, we redesign the attention operator to better handle sparse attention and improve performance during batch decoding. We leverage Triton~\cite{tillet2019triton} to support sparse attention. Instead of a separate mask, our operator adds the Q index as its key input parameter. This Q index is a consolidated list of token indices that require computation in the current step: it includes both the indices of the critical tokens and the indices of the new user-input tokens.
The kernel uses critical tokens' indices to implicitly define the masking logic: it is instructed to ignore the original KVCache positions of the critical tokens within the shared chunk. Instead, it computes attention using the newly computed KVCache for these tokens, which is stored in each query's exclusive page. In our design, each query token performs the standard causal attention calculation only with the valid, non-masked tokens that come before it.

As shown in Figure \ref{fig:q-index}, consider Query 1, which contains 3 tokens and recalls Chunk $b$. The query selects two critical tokens, Token 6 and Token 8, from the chunk. We reserve space for these critical tokens and proceed with reprocessing. During this stage, we compute the KVCache for both the critical tokens and the user-input tokens. The Q index for this process includes tokens 6, 8, 13, 14, and 15. Positions 11 and 12 are reserved in the query's exclusive page to store the new KVCache for the critical tokens (6 and 8). Consequently, the original Tokens 6 and 8 within the shared Chunk $b$ are masked out. After computation, the new KVCache is placed in its reserved space (positions 11 and 12), and decoding continues. This ensures that the KVCache for Chunk $b$ remains intact and available for reuse by other queries.

%% file: chapter/appendix.tex
\appendix
\section{Appendix}
\subsection{Position index recovery and example}
\label{appendix:a}
Here, we prove that Rotary Positional Encoding(RoPE) positional encoding is additive and provide an example to illustrate it.

\textbf{Definition:} Let $k=[k_1,k_2,\cdots,k_d]$ vector to be embedded at position $L$. RoPE encodes the positional information as follows:
\begin{equation}
    \text{RoPE}(k,s) = k\cdot \cos(s \theta)+\text{rotate}(k)\cdot \sin(s \theta)
\label{eq20}
\end{equation}
Here, $\theta=[\theta_1, \theta_1,\theta_2,\theta_2,\cdots,\theta_{\frac{d}{2}},\theta_{\frac{d}{2}}]$ is the rotation frequency parameter related to the dimension, typically defined as:
\begin{equation}
\theta_i = \frac{1}{10000^{\frac{2i}{d}}}, \quad i =  1, \dots, \frac{d}{2};
\end{equation}
$\text{rotate}(k)$ is the operation that swaps the odd and even dimensions of $k$, defined as:
\begin{equation}
\text{rotate}(k)=[-k_2,k_1,-k_4,k_3,\cdots]
\end{equation}
\textbf{Proof:} The first encoding is applied to $k$ at position $s_1$:
\begin{equation}
    k^\prime = k\cdot \cos(s_1 \theta)+\text{rotate}(k)\cdot \sin(s_1 \theta)
\end{equation}
The second encoding is applied to $k^\prime$ at position $s_2$:
\begin{equation}
\begin{aligned}
    \text{RoPE}(k^\prime,s_2) &= k^{\prime} \cdot \cos(s_2 \theta)+\text{rotate}(k^{\prime})\cdot \sin(s_2 \theta) \\
    & = (k\cdot \cos(s_1 \theta)+\text{rotate}(k)\cdot \sin(s_1 \theta)) \cdot \cos(s_2 \theta)\\
    & \quad +\text{rotate}(k\cdot \cos(s_1 \theta) \\
    & \quad +\text{rotate}(k)\cdot \sin(s_1 \theta))\cdot \sin(s_2 \theta) \\
    & = k \cdot \cos(s_1\theta) \cdot \cos(s_2\theta)\\
    &\quad +\text{rotate}(k)\cdot\sin(s_1\theta)\cdot \cos(s_2\theta)\\
    & \quad +\text{rotate}(k)\cdot\cos(s_1\theta)\cdot\sin(s_2\theta) \\
    &  \quad -k\cdot\sin(s_1\theta)\cdot\sin(s_2\theta) \\
    & = k\cdot (\cos(s_1\theta)\cdot\cos(s_2\theta)\\
    &\quad-\sin(s_1\theta)\cdot\sin(s_2\theta)) \\
    & \quad + \text{rotate}(k)\cdot(\sin(s_1\theta)\cdot\cos(s_2\theta)\\
    &\quad +\cos(s_1\theta)\cdot\sin(s_2\theta)) \\
    & = k\cdot\cos((s_1+s_2)\theta)+\text{rotate}(k)\cdot\sin((s_1+s_2)\theta) \\
    & = \text{RoPE}(k,s_1+s_2)
\end{aligned}
\end{equation}
For input $X[1:8]$, where $X[1]$ is system prompt $\mathcal{S}$, $X[2:4]$ is text chunk $C_1$, $X[5:8]$ is text chunk $C_2$. In the preprocessing stage, the two chunks will be contacted by $\mathcal{S}$. The two chunks' KVCache are $KV_1$ and $KV_2$, and the corresponding position index are $[2,3,4]$ and $[2,3,4,5]$. In the KVCache concatenation of Full Reuse, the KCache of the second text chunk needs to adjust its positions from $[2,3,4,5]$ to $[5,6,7,8]$.